\documentclass[10pt,journal,compsoc]{IEEEtran}
\ifCLASSOPTIONcompsoc
  \usepackage[nocompress]{cite}
\else
  \usepackage{cite}
\fi
\ifCLASSINFOpdf
\else
\fi

\usepackage{lineno}
\modulolinenumbers[5]

\usepackage{nicefrac}
\usepackage{caption}
\usepackage[margin=1.5cm]{geometry}
\newcommand{\subparagraph}{}
\usepackage{titlesec}
\usepackage{hyperref}
\usepackage{float}
\usepackage{graphicx}
\graphicspath{ {./images/} }
\usepackage{xcolor}
\usepackage{url}
\usepackage{dsfont}
\usepackage{amsmath,amsfonts,amssymb,amsthm}
\usepackage{mathtools}
\usepackage{commath}
\usepackage[sc,osf]{mathpazo}
\usepackage{subfig}

\theoremstyle{definition}
\newtheorem{definition}{Definition}
\theoremstyle{definition}

\theoremstyle{plain}

\begin{document}


\title{A Matrix Factorization Model for Hellinger-based \\ Trust Management in Social Internet of Things
}

\author{\vspace{-2mm}\text{Soroush~Aalibagi}$^{\dagger}$,~\text{Hamidreza~Mahyar}$^{\dagger}$,~Ali~Movaghar,~and~H.~Eugene~Stanley%
\thanks{$^{\dagger}$ Authors contributed equally}
\thanks{S. Aalibagi and A. Movaghar are with the Department of Computer Engineering, Sharif University of Technology, Iran (e-mail: aalibagi@ce.sharif.edu; movaghar@sharif.edu).}%
\thanks{H. E. Stanley is with the Department of Physics, Boston University, USA (e-mail: hes@bu.edu).}
\thanks{H. Mahyar is with the W Booth School of Engineering Practice and Technology, McMaster University, Canada, and was with the Department of Physics, Boston University, USA (e-mail: mahyarh@mcmaster.ca).}
}

\IEEEtitleabstractindextext{%
\vspace{-2mm}
\begin{abstract}
The Social Internet of Things (SIoT), integration of the Internet of Things and Social Networks paradigms, has been introduced to build a network of smart nodes that are capable of establishing social links. In order to deal with misbehaving service provider nodes, service requestor nodes must evaluate their trustworthiness levels. In this paper, we propose a novel trust management mechanism in the SIoT to predict the most reliable service providers for each service requestor, which leads to reduce the risk of being exposed to malicious nodes. We model the SIoT with a flexible bipartite graph (containing two sets of nodes: service providers and service requestors), then build a social network among the service requestor nodes, using the Hellinger distance. Afterward, we develop a social trust model using nodes' centrality and similarity measures to extract trust behaviors among the social network nodes. Finally, a matrix factorization technique is designed to extract latent features of SIoT nodes, find trustworthy nodes, and mitigate the data sparsity and cold start problems. We analyze the effect of parameters in the proposed trust prediction mechanism on prediction accuracy. The results indicate that feedbacks from the neighboring nodes of a specific service requestor with high Hellinger similarity in our mechanism outperforms the best existing methods. We also show that utilizing the social trust model, which only considers a similarity measure, significantly improves the accuracy of the prediction mechanism. Furthermore, we evaluate the effectiveness of the proposed trust management system through a real-world SIoT use case. Our results demonstrate that the proposed mechanism is resilient to different types of network attacks, and it can accurately find the most proper and trustworthy service provider.
\end{abstract}
\vspace{-2mm}
\begin{IEEEkeywords}
Social Internet of Things, Trust Management, Bipartite Graphs, Matrix Factorization, Hellinger Distance.\vspace{-2mm}
\end{IEEEkeywords}}

\maketitle


\IEEEdisplaynontitleabstractindextext
\IEEEpeerreviewmaketitle

\IEEEraisesectionheading{\section{Introduction}\label{sec:introduction}}\vspace{-1mm}
\IEEEPARstart{T}{he} Internet of Things (IoT) \cite{Burhan_2018} can be considered as the superior technology that facilitates connectivity between heterogeneous physical objects. In order to build a network of objects (a set of smart nodes with the ability to establish social links for information sharing), the combination of IoT and social networking paradigm, \emph{Social Internet of Things (SIoT)}, has been introduced \cite{premarathne2017mag}. Based on \cite{atzori2014smart}, one can observe a generational leap from objects with a certain level of smartness to objects with a concrete social awareness that are able to use environmental consciousness to take appropriate actions; the evolution and progress of the SIoT, containing trillions of objects, cannot be achieved without considering this potential. In SIoT, objects act as autonomous agents. Alongside their individuality, objects can request and provide information and services to each other. The advantages of this convergence are as follows \cite{atzori2012social}:\\
1) A Social IoT guarantees both network navigability, which refers to the effective discovery of objects and services, and network scalability, just like human social networks.\\
2) Levels of trustworthiness could be established by leveraging a degree of interaction among friend autonomous things.\\
3) The previously designed models to study social networks could be extended for reusing in the Social IoT.
\vspace{-3mm}
\subsection{Problem Statement}
\label{problemstatement}
The concept of trust is a longstanding research topic in computer science \cite{artz2007survey}. No categorical consensus on the definition of trust can be found in the scientific literature since trust is a complicated concept. Also, one of the notable challenges is that there is no unified metric or evaluation methodology \cite{sicari2015security, sharma2017cooperative}. In this paper, our trust definition is inspired by the trust notion mentioned in \cite{lin2018clarifying}:

\begin{definition}\vspace{-1mm}
After a service requestor sends out a task for execution, the initiator of the task loses its control of the task; then, the service provider is able to perform its probable malicious animus. Thus, in order to receive the desired service, a service requestor must evaluate the service provider’s competence based on past experiences and decide whether to delegate its task to the service provider. The process of evaluating a service provider’s competence by a service requestor and deciding to delegate a task to that specific service provider is called \textbf{trust}.
\end{definition}

Each SIoT node has an objective trustworthiness value that indicates its reliability and probability of performing attacks that will be discussed later in this section. For convenience, we define \textit{trustor} and \textit{trustee}, as follows:
\vspace{-1mm}
\begin{definition}
\textit{A trustor is a service requestor node in SIoT that has a task to delegate and evaluates the corresponding responses from the service providers.}
\end{definition}
\vspace{-4mm}
\begin{definition}
\textit{A trustee is a service provider node in SIoT that is capable of providing some types of services as the trustor asks and is beyond the trustor’s direct control.}
\end{definition}
\vspace{-1mm}
The nature of trust is context-dependent, i.e., a trustor trusts a trustee in a specific context, but as the context alters, the trustor may decide not to trust that trustee \cite{lin2018clarifying}. The context is related to the characteristics of the service. The trustor expects proper results from the trustee. After getting a response, the trustor measures and rates the trustee's trustworthiness based on the characteristic of the received service and keeps its experience for future decisions.

From the definition of trust between SIoT objects, a malicious node (object) may break the basic functionality of the network by destroying the reputation of well-behaved nodes or increasing the trustworthiness value of malicious ones. In this paper, we focus on the following five prevalent attacks by which a node can violate the existing trust or break the functionality of the devices in the network \cite{chen2016trust, guo2017survey}:

\begin{enumerate}
    \item \textit{Whitewashing Attack}: A malicious node, which has unfavorable reputation, leaves and rejoins the network to avoid the retributions it may encounter due to its poor trustworthiness value. This attack usually happens when the attacker (malicious node) can easily change its identity.
	\item \textit{Self-Promoting Attack}: A malicious node annihilates a well-behaved node's reputation by faking bad experiences about it and thereby intercepts its services.
	\item \textit{Bad-Mouthing Attack}: A malicious node annihilates the reputation of a well-behaved node by faking bad experiences about it, and thereby intercepts its services.
	\item \textit{Ballot Stuffing Attack} or \textit{Good-Mouthing Attack}: A malicious node falsely promotes a misbehaving node (as a well-behaved node) to boost its chance of being selected as a trustee.
	\item \textit{Opportunistic Service Attack}: A malicious node behaves like a non-malicious node in its early appearance to gain a high reputation opportunistically. Then, as soon as earning enough reputation, the node begins its malicious behaviors, like previous attacks.
\end{enumerate}

Trust management is a mechanism to predict the most reliable trustee for a certain trustor. Trust management allows SIoT objects to eliminate the risk of exposure to malicious nodes. A practical and effective trust management mechanism must fulfill the following requirements:

\begin{itemize}
  \item \textit{Resistance to attack}: A trust management mechanism should provide resiliency against related attacks.
  \item \textit{Overcoming resource constraints}: In IoT, objects are often very constrained in terms of memory, power supply, and processing power \cite{al2015internet}. Therefore, strong security measures, like heavy cryptography, are not applicable \cite{sha2018security}, and particular novel solutions are required \cite{esposito2018distributed}.
  \item \textit{Locality}: A trust management method must not critically depend on a remote central node, which restricts the scalability and induces harmful effects. For instance, a pitfall in centralized trust management can cause the failure of the whole system \cite{kim2017securing}. Moreover, the ability to scale on large IoT networks makes the system viable \cite{AIREHROUR2019860}.
  \item \textit{Robustness against data sparsity}: Trust management systems should not suffer from the data sparsity problem and the cold start problem, which is an inability of the system to properly predict trust values due to the lack of enough data for recently joined users.
\end{itemize}

Until now, we have described the concept of trust and the requirements for a trust management system in SIoT. Next, we explain the motivation and importance of trust management in SIoT.

\subsection{Motivation}
\label{motivation}\vspace{-1mm}
Trust is an emerging topic, and a fundamental issue since several devices with various behaviors characterize the SIoT environment \cite{sicari2015security, butun2019security}.
The motivation for proposing an efficient trust management method for SIoT is discussed in \cite{chen2016trust} and \cite{9026905}. To the best of our knowledge, previously proposed trust management systems are not efficient enough to be deployed in a real-world scenario and cannot fulfill all the requirements discussed in the previous subsection. We will discuss related work in the next section.

For instance, we can imagine SIoT sensors (trustors) in traffic intersections in a smart city. Such SIoT devices can connect to smart cars (trustees) passing the intersection and push an alarm notification on blind or deaf pedestrians' smartphones passing the intersection. To avoid sending false negative and positive alarms, intersection sensors need a trust management system to determine trustees' trustworthiness. Moreover, such trust management methods should \textbf{overcome data sparsity} problems since new cars, with which the intersection sensor does not have any previous experience, appear in the network and cannot be fully trusted. Further, the trust management method must \textbf{overcome resource constraints} since it most probably uses solar power, which is limited in cloudy weather.

Another scenario is the IoT-based weather reporting system designed to facilitate the live reporting of weather parameters using individuals' smartphones. For example, one's smartphone that wants to check Paris's weather needs to run a trust management mechanism to receive weather parameters from accurate and trustworthy IoT nodes in Paris. Accordingly, as the number of IoT nodes grows, the trust management method has to support \textbf{scalability}. Wang et al. \cite{9026905} describe another scenario showing that trust management systems have to provide \textbf{resiliency against the attacks} mentioned earlier.

However, there is no sufficient work on trust management in SIoT \cite{sharma2017cooperative, Burhan_2018}. Consequently, we aim to present a novel trust management mechanism to predict the level of trust between the SIoT nodes, using social relations between devices. Section \ref{experimentalevaluation} shows that our method is accurate yet resilient to different attacks and satisfies the aforementioned requirements.


\section{Related Work}
\label{relatedwork}
There exist very limited relevant works on trust management in SIoT, particularly considering misbehaving users who attack the well-behaved users through their possessed SIoT devices \cite{nitti2014trustworthiness, chen2016trust, sharma2017cooperative, guo2017survey}. 
Chen et al. \cite{chen2016trust} developed a scalable and adaptive trust management protocol for SIoT systems and tried to reduce the probability of being attacked by dynamically changing devices' configuration. Their model is user-based, which means the network nodes have social relations through their owners' social network. Hence, the social relations between SIoT objects are constrained to the relations between the owners. The other weakness of this model is that the authors divided SIoT devices into two inflexible types (\textit{i.e.,} devices and owners), such that trustees should be selected from the pre-defined devices, and trustors are only selected from the owners. Finally, they addressed the resource constraint of IoT devices meticulously and proposed storage management to overcome this issue.

Nizamkari \cite{nizamkari2017graph} followed the same idea as Chen et al. \cite{chen2016trust}, and in their work, trustors use their own experiences and their friends' experiences to evaluate the trustworthiness of trustees. However, the most important difference between them is that, in the former, if the trustor's friends do not have appropriate requested experiences, the trustor inquires its friends of friends. The influence of recommendation from friends of friends could be obtained from nodes similarity or the network structure. Nevertheless, there are two inabilities to tackle the prediction issue. First, their proposed recommender cannot predict the rating for an object that has not been rated yet. Second, the author did not propose any solution for the situation when searching nodes for finding friends of friends (experience) terminates in identical results.

Kantarci et al. \cite{kantarci2014trustworthy} studied a cloud-centric IoT, which is called crowd computing. In their framework, mobile sensors are used as IoT devices that reside in the cloud. In this scenario, a user issues a task, and a crowd management authority candidates some of the social network users for assigning the task. The most significant weaknesses of this framework are the need for a central node to manage the issues, store users' reputation, and process a large amount of data per issue. Today, accepting a central node for processing and managing all the tasks is not practical \cite{nizamkari2017graph}, because of the scalability issues as the number of IoT devices is tremendously growing. Nonetheless, they could degrade the impact of malicious users significantly.

Mendoza and Kleinschmidt \cite{mendoza2018distributed} used the experiences of trustors' neighbors and the quality of trustees' services to evaluate the trust between trustors and trustees. However, each trustor must store a table containing all other nodes and their experiences. This trust management model can detect malicious behaviors, but storing such a large table is non-practical for lightweight IoT objects.

Chen et al. \cite{chen2011trm} proposed a trust management model based on a fuzzy reputation to evaluate trust in IoT systems. For trust composition, they consider only QoS trust metrics such as end-to-end packet forwarding ratio, energy consumption, and packet delivery ratio. Moreover, their trust management model considers only wireless sensors, which is specific to the IoT environment. Furthermore, they did not consider the social relationship between the objects.

Sharma et al. \cite{sharma2017cooperative} proposed a novel solution for the maintenance of trust and preservation of privacy rules in SIoT in the form of a lightweight query mechanism with the help of fission computing. Although the authors asserted that the trust would be provided without sanctioning adversaries, \cite{imran2019enabling} claimed that users might expose selfish behaviors in relaying data to others due to limited resources or social objectives. Furthermore, demanding for 5G infrastructure and an administrative center for queries are other weaknesses of this mechanism. 

Yu et al. \cite{yu2018iotchain} suggested the blockchain for data management to provide end-to-end trust and remove the trusted third-party in decentralized IoT systems. However, there is a trade-off between scalability and privacy in their proposed mechanism. Hence, it could not afford both parameters, simultaneously. Moreover, the authors ignored the fundamental constraints of IoT objects.		

Due to the problems of fully-distributed or centralized trust management systems, Kim \cite{kim2017securing} proposed an authentication and authorization infrastructure for IoT to be locally centralized and globally distributed. This solution scales well and utilizes edge devices. However, it ignores the dynamicity and sociality of the network devices.

Chen et al. \cite{chen2016trustt} proposed a scalable and adaptive trust management protocol in SOA-based SIoT systems, which is distributed and utilizes users' feedbacks using similarity rating of friendship, social contact, and community of interest relationships. Their protocol takes some SIoT constraints into account, such as limited storage and computing capacity of devices by storing trust information only for a limited set of nodes which leads to less trust updates. However, considering only a limited set of nodes is not fundamentally a scalable solution.


Awan et al. \cite{awan2019holitrust} proposed a trust management system to manage trust during cross-domain communications in the IoT. Their system relies on central authorities to handle all domains, computes and stores the trust of domains, and issues certificates; thus, the system's scalability will not be guaranteed. Further, social relationships between the objects and misbehaving users who carry out the attacks, as mentioned earlier, are not considered in their system. Nevertheless, they addressed the IoT devices' resource constraint meticulously and reduced the computational weight on IoT nodes.

Alemneh et al. \cite{alemneh2019two} proposed one of the first bidirectional trust management systems for fog computing, which is a promising technology to realize the global-scale IoT. Their solution lets service requestors verify service providers' reliability and helps service providers investigate the trustworthiness levels of service requestors. Moreover, their system is resilient to various types of misbehaving nodes. However, in their trust calculation, the considered social relationship is restricted to ownership relations.

Overall, a few previous works on trust management of SIoT test their proposed solutions against different attacks, which have been mentioned earlier in Section \ref{motivation}. In this paper, we propose a novel trust management mechanism in SIoT and analyze its performance in the presence of malicious nodes.

\vspace{-3mm}
\section{Proposed Method}
\label{proposedmethod}
In this section, we (1) introduce a bipartite graph model for SIoT (Section \ref{architecture}); (2) build a social network among trustors using Hellinger distance (Section \ref{socialnetworkoftrustors}); (3) introduce a new social trust model for trustors based on the constructed social network (Section \ref{socialtrustmodel}); (4) introduce a new matrix factorization-based mechanism for trustors to help them find the most trustworthy trustee (Section \ref{predictionmechanism}); (5) discuss (in Section \ref{howmymethodaddressestherequirements}) how the proposed model fulfills the required obligations mentioned in section \ref{problemstatement}; (6) analyze the complexity of our model in Section \ref{timecomplexity}, eventually.

\begin{figure}[]
\vspace{-4mm}
\centering
\captionsetup{justification=centering}
\includegraphics[width=0.45\textwidth]{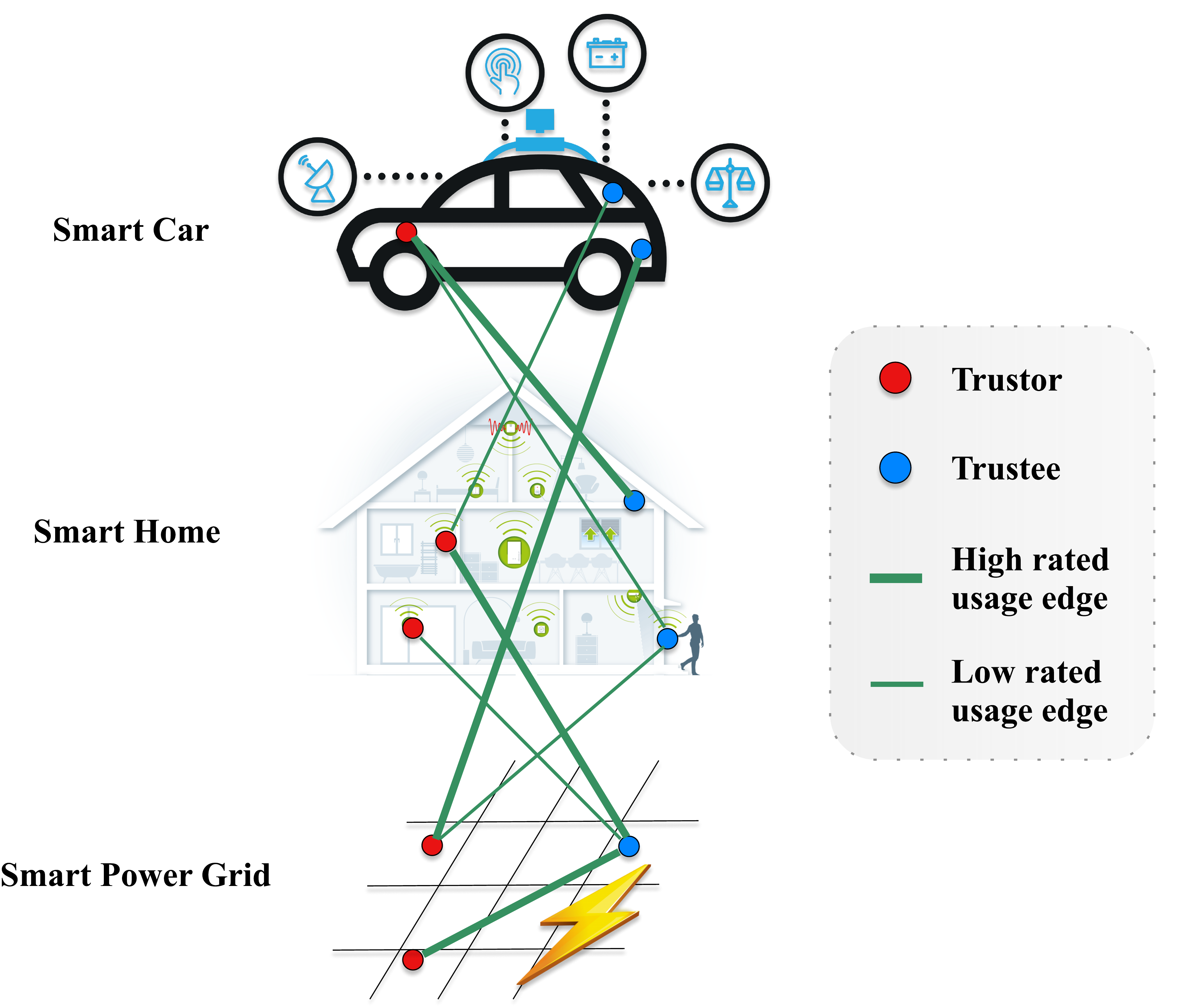}
\vspace{-5pt}
\caption{\small A bipartite graph indicating a part of an IoT system.}
\label{fig:sec}
\end{figure}

\textfloatsep 2.5mm
\vspace{-3mm}
\subsection{SIoT Bipartite Graph Model}
\label{architecture}
Based on the applications of SIoT objects, a well-known architecture for SIoT would be service-oriented \cite{chen2016trust}. Each device in the system can play the role of trustor, trustee, or both \cite{chen2016trust}. Hence, we believe that bipartite graphs are suitable models for representing the social internet of things (Figure \ref{fig:sec}). Without loss of generality, one can assume that there is a limited number of service types that could be performed among SIoT objects. For each service type, we can consider a separate bipartite network with two sets of nodes, trustors $U$ and trustees $V$. Considering a specific bipartite network, trustor $u \in U$ has a directed link (edge) to trustee $v \in V$ if and only if trustor $u$ has used the services provided by trustee $v$ at least once. Each edge in the bipartite network has a weight in the interval $(0, 1]$ that indicates the trustor's trust experience of using the trustee's service.  

Let $G=(U,V,E)$ be a bipartite graph with two sets of nodes, trustors $U=\{u_1,u_2,...,u_n\}$ and trustees $V=\{v_1,v_2,...,v_m\}$, and $E=\{(u,v)| u\in U, v\in V\}$ represents the weighted edges from trustors to trustees. Bi-adjacency matrix $B$ of the graph is a matrix of size $n \times m$, wherein $b_{i,j} \neq 0$ if and only if $(u_i, v_j) \in E$. Row $i$ in the bi-adjacency matrix is a vector corresponding to trustor $u_i$ and represents the experience rate of this trustor to any trustee.

\vspace{-2mm}
\subsection{Social Network of Trustors} 
\label{socialnetworkoftrustors}
In this section, we extract implicit social relations between trustors from the aforementioned bipartite graph based on the trustors' past experiences. The extracted social relations can indicate behavioral \textit{trust similarity} among trustors in the network. We generate a social network of trustors using a distance metric, since the similarity measures are, in some sense, the inverse of the distance metrics \cite{taheri2017extracting, guo2013novel}.

\textit{Hellinger distance} (aka Bhattacharyya distance) is a type of \textit{f-divergence} metric introduced by Ernst Hellinger \cite{nikulin2001hellinger}. We have chosen Hellinger distance as a distance metric, because: (1) as the trust concept is inherently non-deterministic, we need a statistical metric to measure distances between the nodes; (2) Hellinger distance satisfies the \textit{triangle inequality}, so that the differences between trustors will be properly demonstrated; (3) it holds \textit{symmetry} and \textit{positive definite} properties that are essential for a well-defined distance metric \cite{taheri2017extracting, taheri2017hellrank}.

To measure the similarity between each pair of trustors, we apply Hellinger distance to the degree distribution of their neighbors.
Let $L_u=\{l_k\}_{k\in [\,d]\,}$ be the probability distribution over all neighbors of trustor node $u$, where $l_k$ is the number of neighbors of $u$ with degree of exactly $k$, and $d$ is the greatest degree of trustees in the network.
The following equation shows the Hellinger distance between two trustors $u_{i}$ and $u_{j}$ as described in \cite{taheri2017extracting}: \vspace{-2mm}
\begin{equation}
Hell(u_i,u_j)=\frac{1}{\sqrt{2}} \norm{L_{u_i}-L_{u_j}}
\label{eqn:helldiss}
\end{equation} 
Now, we have a $n \times n$ distance matrix, where $n$ is the number of trustors in the bipartite network. By considering an appropriate \textit{threshold}, a social relation between any pair of trustors could be formed, based on how close nodes are to each other. As a result, a new social network of trustors is created. The network can be displayed by adjacency matrix $A_{n \times n}$ as: \vspace{-1mm}
\[
    A_{i,j}= 
\begin{cases}
    1		& Hell(u_i,u_j)< threshold \\
    0		& \text{otherwise}
\end{cases}
\]
\vspace{-5mm}

\subsection{Social Trust Model}
\label{socialtrustmodel}
In this subsection, we aim to develop a trust model which expresses how similar two trustors are, based on their trust patterns. For this purpose, past experiences of trustors, the constructed social network in the previous section, and various similarity and centrality measures are employed.
\vspace{-2mm}
\subsubsection{Similarity}
\label{similarity}
The similarity between network nodes (trustors) is one of the essential factors that adheres to their trust patterns \cite{golbeck2009trust, ye2017collaborative, siegrist2000salient}. Different similarity metrics are used in order to extract trust patterns among users of the network.

\textbf{Bayesian Similarity.}
Cosine similarity (COS) and Pearson correlation coefficient (PCC) are widely used to determine the degree of similarity between nodes in a network. However, they suffer from substantial shortcomings, as described in \cite{guo2013novel} and \cite{ahn2008new}. Accordingly, we chose the Bayesian similarity (BS) metric, which is based on the Dirichlet distribution \cite{guo2013novel}. Even though this similarity is a rating-based measure, it does not suffer from the data sparsity problem, or many other problems pointed out in \cite{wang2014vsrank}. BS does not consider mutually rated trustees; instead, it utilizes the preferences of trustors by considering both direction (rating distances) and length (rating amounts) of rating/experience vectors.
The Bayesian similarity between trustors $u_i$ and $u_j$ could be calculated from the following equation, as demonstrated in \cite{ahn2008new}:
\begin{equation}
BS(u_i,u_j)=max(BS'_{u_i,u_j} - BS''_{u_i,u_j} - \delta,0)
\end{equation}
In this relation, $BS'_{u_i,u_j}$ is called \textit{overall similarity}. The Bayesian similarity is computed by removing the chance correlation $BS''_{u_i,u_j}$ and user bias $\delta$ from the overall similarity. Besides, the similarity originated from the data sparsity problem would be resolved by subtracting the chance correlation \cite{taheri2017extracting, guo2013novel}.

\textbf{Hellinger Similarity.}
For Hellinger similarity, we use the Hellinger distance discussed in Equation (\ref{eqn:helldiss}). This measure is analogously rating-based but tolerates the problems mentioned earlier.

\textbf{Connection Similarity.}
The connection similarity takes advantage of connections in the proposed social network of trustors to derive similarity between trustors.
Let $F(u_i)$ be the list of friends for trustor $u_i$. For each pair of nodes, the connection similarity is computed with the following equation \cite{davoudi2016modeling, davoudi2018social}: $conn(u_i,u_j)= \nicefrac{\displaystyle \mathopen| F(u_i) \cap F(u_j) \mathclose|}{\displaystyle \mathopen| F(u_i) \mathclose|} $ 
\vspace{-2mm}
\subsubsection{Centrality}
\label{centrality}

Another criterion in social networks which has a significant impact on nodes' trustworthiness is centrality \cite{mahyar2018dicenod, mahyar2019csclose}. Hence, the following centrality measures are considered to be used in order to discover the trustors' behavior.

\textbf{Degree Centrality.}
Degree centrality (basically) indicates the importance of nodes in a social network \cite{Mahyar2015TopK}, relative to the number of edges (links) incident upon a node. The degree centrality of a trustor $u_i$ can be formalized as $ \displaystyle deg(u_i) = \sum\nolimits_{\forall j,j \ne i} A_{i,j} $ , where $A_{i,j}$ is an element of the adjacency matrix of trustors' social network that indicates the connection between trustors' $u_i$ and $u_j$.

\textbf{Betweenness-Local Clustering Centrality (BLC).}
Since we need a centrality measure to evaluate node's influence on its neighborhood, we employ betweenness-local clustering (BLC) centrality as described in \cite{tong2014novel}. Even though the betweenness centrality \cite{mahyar2018compressive} partially describes the importance of nodes, as it is a global evaluation parameter, it cannot precisely present the relative influence of nodes in a local environment, especially in large-scale complex networks \cite{tong2014novel}. The importance of node $u_i$ would be attained more accurately with the combination of betweenness centrality and the local clustering coefficient \cite{tong2014novel, Mahyar2015CScomdet}, as $ \displaystyle BLC(u_i)= \nicefrac{\displaystyle BC_{u_i}}{\displaystyle CC_{u_i}} $, where $BC_{u_i}$ is betweenness centrality of node $u_i$, and $CC_{u_i}$ denotes its local clustering coefficient.

\vspace{-2mm}
\subsubsection{Trust Pattern Similarity}
\label{trustpatternsimilarity}
In order to achieve the trust pattern similarity of trustors, we utilize the combination of similarity and centrality to create a new measure to evaluate how much trustors $u_i$ and $u_j$ have similar behavior in trusting the trustees \cite{davoudi2018social, davoudi2016modeling}:
\vspace{-1mm}
\begin{equation}
\Gamma(u_i,u_j)=\beta \frac{Sim(u_i,u_j)}{\displaystyle \sum_{u_k|A_{u_i,u_k}=1} Sim(u_i,u_k)} +(1-\beta)\frac{Cen(u_j)}{\displaystyle \sum_{u_k|A_{u_i,u_k}=1}Cen(u_k)}
\label{eqn:linfun}
\end{equation}

where parameter $\beta$ indicates the amount of contribution of similarity and centrality. Indeed, $\Gamma(u_i, u_j)$ constructs a similarity matrix based on the trustors' trust pattern.
\vspace{-2mm}
\subsection{A Prediction Mechanism via Matrix Factorization}
\label{predictionmechanism}
Matrix factorization has become a powerful technique in bipartite network analysis \cite{koren2009matrix} and is a practical technique to engage with trust relations \cite{ma2011recommender, yang2017social}. Here, we apply this technique to predict the trustees' trustworthiness values in SIoT; then, introduce a mechanism that helps trustors to find the most trustworthy trustee in order to entrust their tasks.

There are three influential factors to consider when a trustor wants to select a particular trustee to dispatch its task: (1) features of the trustor, (2) features of the trustee, and (3) the previous (trust) experiences between them. It is challenging to extract, maintain, and update the features of nodes. Besides, there are mostly very few experiences available, particularly when the nodes are newcomers to the network. Therefore, we chose a matrix factorization model \cite{ma2011recommender, davoudi2018social, davoudi2016modeling} to extract latent features and mitigate the data sparsity issue. The premise behind a latent feature extraction model is that there is only a small number of (latent) features influencing the nodes' behavior. To the best of our knowledge, this is the first trust management mechanism in SIoT that employs a matrix factorization model.

In order to learn and generate two $L-$dimensional latent feature representations of trustors $S$ and trustees $R$ matrices, the bi-adjacency matrix $B$ is factorized. Thus, each column of $S \in \mathds{R}^{L \times n}$ and $R \in \mathds{R}^{L \times m}$ respectively displays an $L-$dimensional trustor and trustee latent feature vector. Due to the fact that trustors only have experiences with a limited number of trustees, the bi-adjacency matrix $B$ is highly sparse.
Here, the low-rank matrix factorization approach seeks to approximate the bi-adjacency matrix $B$ by multiplying the two $L-$dimensional factors $S$ and $R$, i.e., $B \approx S \times R $.

Generally, the following cost function is utilized in matrix factorization problems to reconstruct and predict the missing values of bi-adjacency matrix $B$:
\vspace{-2mm}
\begin{equation}
\displaystyle \mathcal{L}(S,R,B)= \frac{1}{2} \| B-S^TR \|_F^2
\label{eqn:firstcost}
\vspace{-2mm}
\end{equation}
where $\| . \|_F^2$ implies the \textit{Frobenius} norm. As mentioned earlier, $B$ is highly sparse, so we only should factorize the existing values. Thus, the above cost function can be reduced to:
\vspace{-2mm}
\begin{equation}
\displaystyle \mathcal{L}(S,R,B)= \frac{1}{2} \sum_{i=1}^n \sum_{j=1}^m \mathbb{I}_{ij} (B_{ij}-S_i^TR_j)^2
\vspace{-2mm}
\end{equation}
where, $\mathbb{I}_{ij}$ is an indicator function that takes value $1$ if $B_{ij}$ exists and $0$ otherwise.

Users in social networks mostly trust their friends \cite{ma2011learning}; consequently, we assume that objects in SIoT trust their social network friends, which have been specified in Section \ref{socialnetworkoftrustors}. Therefore, depending on how much a trustor object is similar to its friends based on the trusting behavior function $\Gamma$ in Equation (\ref{eqn:linfun}), reconstruction of matrix $B$ relies on both the trustor's features and its friends' features. So, we can redefine the cost function as:
\vspace{-4mm}
\begin{multline}
 \displaystyle \mathcal{L}(S,R,B)= \\
 \frac{1}{2} \sum_{i=1}^n \sum_{j=1}^m \mathbb{I}_{ij} \Big( B_{ij}- g \big( \alpha S_i^TR_j + (1-\alpha)\sum_{k \in F(i)} \Gamma_{ik} S^T_kR_j \big) \Big) ^2 
\label{eqn:rabete_dah}
\end{multline}

where $\alpha$ balances between the two mentioned factors and $k \in F(i)$ shows the social network friends (neighbors) of trustor $u_i$.
The argument that passed to $g(.)$ is employed to predict missing values of $B$, which may exceed the valid range $(0, 1]$; hence, it is mapped through a nonlinear logistic function $g(x)=1/(1+exp(-x))$ to rebound it to the valid range.

One can also add a regularization term to the cost function to avoid the over-fitting issue. Finally, the sum-of-squared-errors cost function with the quadratic regularization terms could be defined as:
\vspace{-3mm}
\begin{multline}
 \displaystyle \mathcal{L}(S,R,B) = \\ \frac{1}{2} \sum_{i=1}^n \sum_{j=1}^m \mathbb{I}_{ij} \Big( B_{ij}- g \big( \alpha S_i^TR_j + (1-\alpha)\sum_{k \in F(i)} \Gamma_{ik} S^T_kR_j \big) \Big) ^2 \\ + \frac{\lambda_{S}}{2} \| S \|_F^2 + \frac{\lambda_{R}}{2} \| R \|_F^2 
\label{eqn:yazda}
\end{multline}
where hyper-parameters $\lambda_{S}, \lambda_{R} > 0$ are $S$ and $R$ latent variance ratios. As finding the global optimum is often difficult, Stochastic Gradient Descent (SGD) could be applied several times to locate the best local minimum \cite{ma2011recommender}. Also, this cost function has an attractive probabilistic interpretation with Gaussian observation noise (for more detail, see \cite{mnih2008probabilistic}).

\begin{figure*}[]
\vspace{-5mm}
\centering
\captionsetup{justification=centering}
\includegraphics[width=0.8\textwidth]{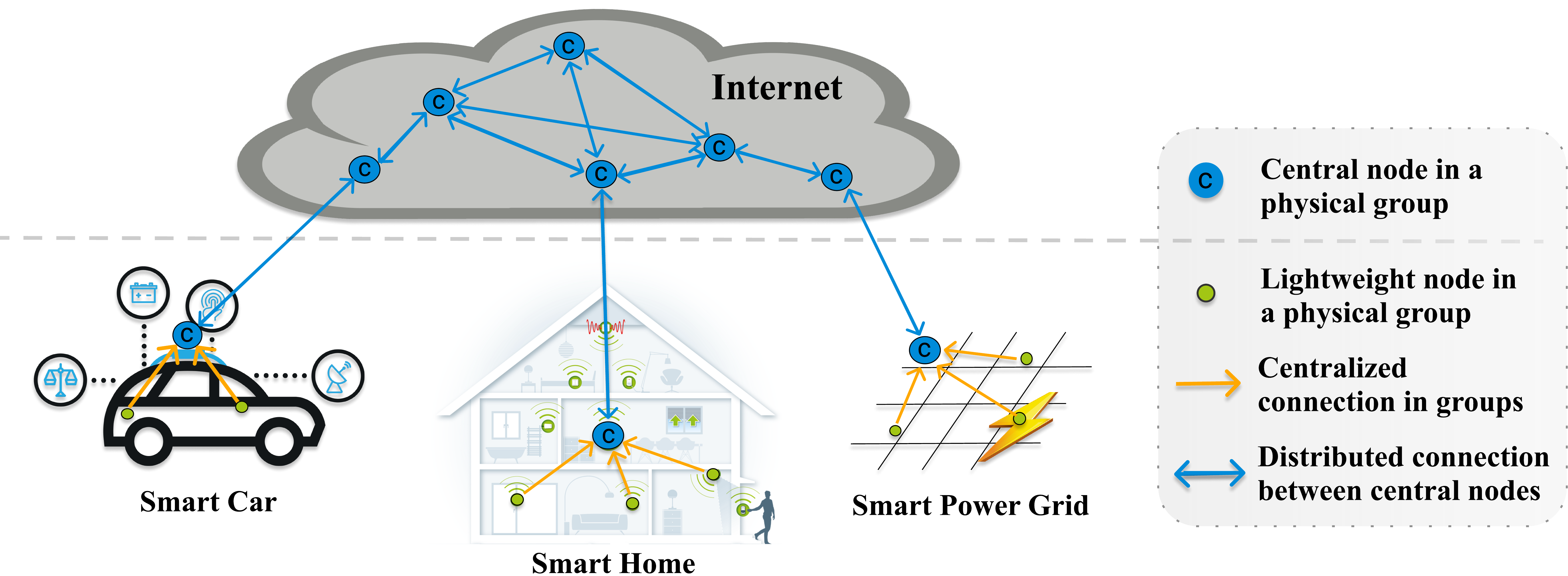}
\vspace{-1.5mm}\caption{\small Illustration of the \emph{locally centralized, globally distributed} architecture of a SIoT system: figures under the dashed line depict three locally centralized groups, and the cloud on top shows the Internet that includes a globally distributed network of central nodes of physical groups.}
\vspace{-4mm}
\label{fig:first}
\end{figure*}

Ultimately, a trustor node $u_i$ can acquire the two latent feature matrices $S$ and $R$, by having the bi-adjacency matrix $B$ and using the cost function above; then, it can utilize the following equation in order to reproduce $B$, i.e., $ \widehat{B} =g(S^TR) $,
where $g$ is the non-linear logistic function.

Afterward, the trustor $u_i$ can extract the $i$-th row of $\widehat{B}$ and sort it to obtain a vector of trustees ordered by their trustworthiness value. Consequently, a trustor can select the most trustworthy trustee to dispatch its task. In this way, a trustor can predict which service provider (trustee) is the most reliable one to connect.
\vspace{-2mm}
\subsection{Requirements Satisfaction by Our Method}
\label{howmymethodaddressestherequirements}
Here, we demonstrate how the proposed trust management mechanism meets the requirements discussed in Section \ref{problemstatement}.
We use the \emph{locally centralized, globally distributed} architecture
to show how a trustor can predict the most trustworthy trustee for its requested service in a scalable and distributed manner. As depicted in Figure \ref{fig:first}, we assume each SIoT object is a member of a physical group with a central node. For example, considering a smart home as a group, it contains lightweight objects such as smart light and smart thermostat and non-lightweight objects like smart TV and Google Home assistant. Thus, these groups have at least one non-lightweight node that could be selected as the central node of its group. Moreover, all the SIoT objects in a group have the same behavior as their owner does since each group is owned by just one individual or organization. So, not all trustor nodes need to perform the trust management mechanism themselves. Instead, whenever nodes of a group need to select an external service provider, the group's central node collects prior experiences from the other nodes of the group and also from other groups to perform the matrix factorization. Then, the central node gives each node its related row from the reconstructed bi-adjacency matrix $\widehat{B}$. In this way, each group is locally centralized, but the whole system is globally distributed, and the lightweight SIoT devices do not have to factorize any matrix, which is a computationally expensive task or save any data more than their own experiences. Therefore, the whole system is \textit{not dependent on a remote central node} and \textit{overcomes SIoT devices' constraints}.

As mentioned earlier, the matrix factorization mechanism intrinsically \textit{alleviates the data sparsity problem}. Further, we assume that each node in SIoT has an identification number (like MAC address); thus, its trust information could be saved along with its identifier. Consequently, if a node decides to leave and rejoin the network, its trust data will not be lost. Hence, our trust management protocol deals with nodes that perform a \textit{whitewashing attack}. Furthermore, it is obvious that selecting an appropriate threshold for \textit{Hellinger distance} leads to the construction of a suitable network between trustors. This social network specifies each node's friends in such a way that bad behavioral nodes leave the network, and none of the mentioned attacks (in section \ref{problemstatement}) could affect our prediction protocol. We will discuss the malicious nodes performing those attacks in detail in the next section.

\vspace{-1mm}
\subsection{Time Complexity}
\label{timecomplexity}

To obtain the time complexity of our method, we investigate the complexity of each part separately. To construct the bipartite SIoT graph, it takes $O(n^2 \Delta+l)$ time to generate the Hellinger distance matrix, where $n$ is the number of trustors, $\Delta$ is the highest degree of trustees in the network, and $l$ is the number of links in the bipartite network. In order to calculate the trust pattern similarity between trustors to develop a social trust model, we employ various combinations of similarity and centrality measures. The most time-consuming combination takes $O(n \times e)$, which is related to the BLC centrality calculation, where $e$ denotes the number of edges in the constructed social network of trustors. Finally, Computing matrix factorization with SGD requires $O(z \times d \times L)$ time that $z$ is the number of non-zero entries of matrix $B$, $d$ is the average degree in the social network of trustors, and $L$ is the number of latent features, which is a small constant here. Hence, the overall time complexity can be expressed as $O(n^2 \Delta+l+n \times e+z \times d \times L)$.

%
%
\vspace{-1mm}
\section{Experimental Evaluation}
\label{experimentalevaluation}
In this section, we evaluate the performance of the proposed trust management system in three different scenarios. First, we investigate the quality of the matrix factorization mechanism. Then, we follow the evaluation procedure outlined in \cite{chen2016trust} to assess our trust management model's performance. Finally, we apply our trust management mechanism to a real-world SIoT use case to exhibit the utility of our protocol.

\vspace{-2mm}
\subsection{Accuracy of the Matrix Factorization Mechanism}
\label{validityandaccuracyofmf}
To test the accuracy and quality of the matrix factorization model, we have performed several experiments on the Epinions dataset\footnote{\href{http://www.cse.msu.edu/\~tangjili/trust.html}{www.cse.msu.edu/~tangjili/trust.html}}. We also compare the proposed method with the best existing trust prediction methods.

\subsubsection{Dataset}
\label{datasets}
$\mathtt{Epinions.com}$ is a well-known review website where users can review products and assign them integer ratings from 1 to 5. 
We employ our Hellinger-based technique to construct a social network, using this rating matrix. This dataset consists of 922,267 ratings given by 22,166 users to 296,277 items, which leads to an extremely sparse rating matrix with a density percentage of 0.014.
On average, each user rates 41.607 times in her era.
\vspace{-2mm}
\subsubsection{Settings \& Metrics}
\label{settingsmetrics}
We used predictive accuracy and classification accuracy measures for evaluating the proposed matrix factorization method. Accuracy of a prediction system measures the closeness of the method’s predicted ratings to the ground-truth (actual) user ratings \cite{herlocker2004evaluating}.


\textit{Root Mean Squared Error (RMSE)} is a popular predictive accuracy metric. It is the square root of the average of squared differences between the true and the predicted ratings, as:
\vspace{-1mm}
\begin{equation}
\displaystyle RMSE =\sqrt{ \frac{ \sum_{i,j} |B_{i,j}^{pre} - B_{i,j}^{act}|^2}{N} }
\label{enq:rmse}
\vspace{-1mm}
\end{equation}

Where $N$ is the number of nonzero elements in matrix $B^{act}$. Although both MAE and RMSE express average prediction error, RMSE can be more relevant in cases that large errors are intolerable.

\textit{Coverage} is the percentage of ratings that our method is able to generate predictions. Systems with higher coverage are more advantageous since there are more decisions they are able to help with \cite{herlocker2004evaluating}. Coverage can be defined as:
\vspace{-1mm}
\begin{equation}
Coverage = \frac{\text{\# ratings that system can make prediction}}{\text{\# available ratings (items) to predict}}
\vspace{-1mm}
\end{equation}

\textit{Precision}, within this context, is associated with the normalized form of RMSE and obtained as follows \cite{jamali2009trustwalker}:
\vspace{-1mm}
\begin{equation}
Precision = 1- \frac{RMSE}{RMSE_{max}}
\vspace{-1mm}
\end{equation}
where $\text{RMSE}_{max}$ is the maximum possible value of the RMSE.

\textit{F-measure} is a harmonic mean of precision and coverage to consider both metrics into a single evaluation metric. It is defined as \cite{jamali2009trustwalker}:
\vspace{-2mm}
\begin{equation}
F\text{-}measure = \frac{2 \times Precision \times Coverage}{Precision + Coverage}
\vspace{-1mm}
\end{equation}


Our algorithm's best desired setting can be achieved when we set the size of the latent features $L$ to $4$ and the parameters $\alpha$ in Equation (\ref{eqn:rabete_dah}) and $\beta$ in Equation (\ref{eqn:linfun}) to $0.4$ and $1$, respectively. We also use $75$ percent of the Epinions data as the training set and the rest for the test set. The reason for picking such values for the aforementioned parameters will be discussed later in the subsequent sections. The hyper-parameters $\lambda_{S}$ and $\lambda_{R}$, in Equation (\ref{eqn:yazda}), are also set to $0.001$, as suggested by \cite{ma2011recommender}.

\vspace{-2mm}
\subsubsection{Evaluation Results} 
\label{results}
\begin{figure}[]
\vspace{-2mm}
\centering
\captionsetup{justification=centering}
\includegraphics[width=0.48\textwidth]{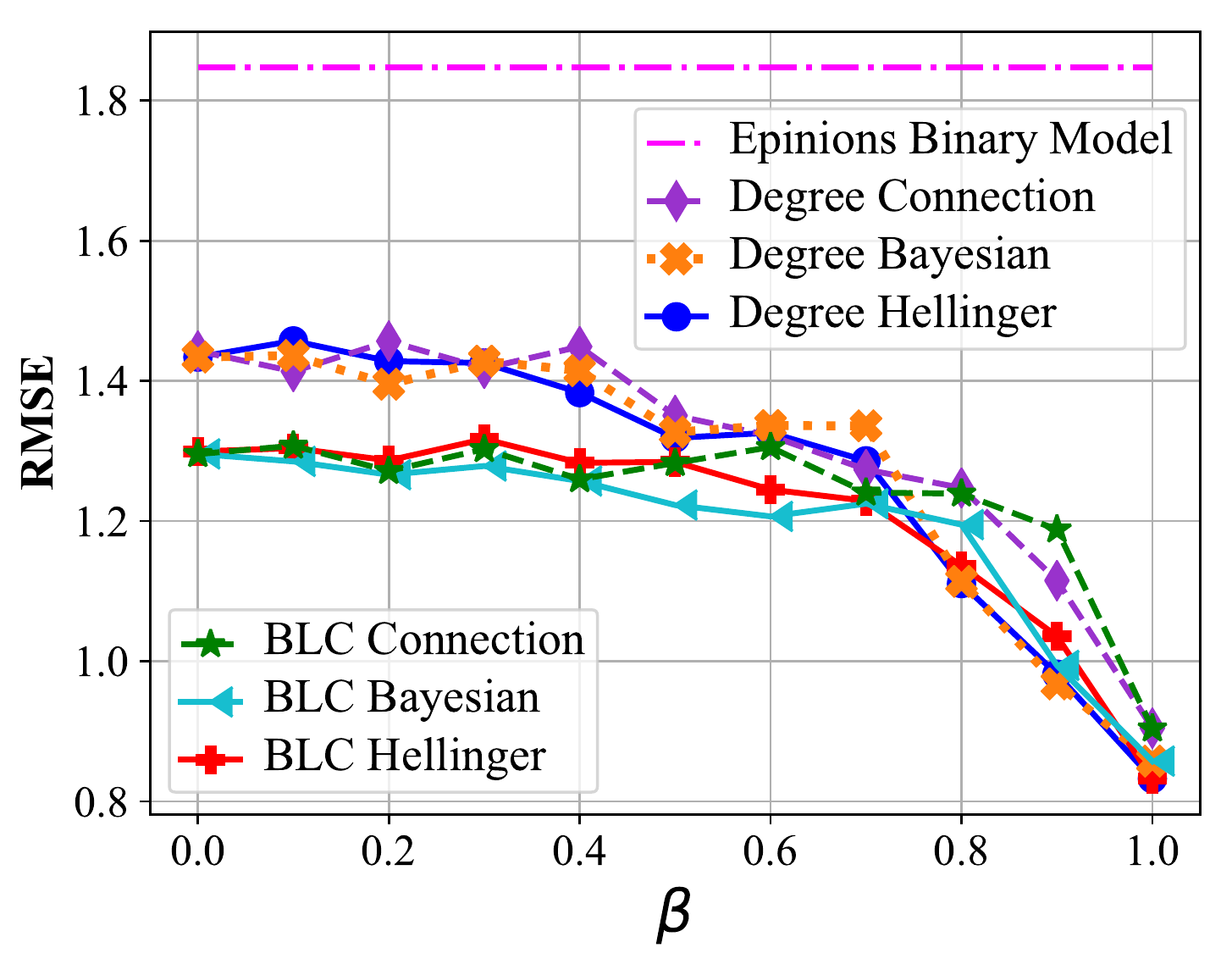}
\vspace{-5mm}
\caption{\small RMSE of our prediction mechanism using different combination of centrality and similarity measures, for different values of $\beta$. Also, RMSE comparison of Epinions-based binary trust model and our prediction protocol.}
\label{fig:all_rmse}
\vspace{-1mm}
\end{figure}

According to Equation (\ref{eqn:linfun}), the impact of trustor's friends is introduced as the linear combination of the friends' centrality and the similarity between them with the weighting factor $\beta$. As shown in Figure \ref{fig:all_rmse}, we analyze how the changes of $\beta$ can affect the prediction accuracy in terms of RMSE measure. Also, we introduce a binary trust model and assess its RMSE.

The binary trust model is a model that the similarity and centrality of nodes do not affect the trust pattern similarity between nodes, and just the friendship between nodes in the social network is considered. That means, for the binary trust model, $\Gamma$ is equal to $1$ in Equation (\ref{eqn:rabete_dah}). The straight magenta line in Figure \ref{fig:all_rmse} shows the RMSE of binary trust model using the Epinions dataset users network. As depicted in Figure \ref{fig:all_rmse}, our prediction mechanism based on trust pattern similarity, regardless of which centrality and similarity measures it uses, outperforms the binary trust model. This observation specifies that the social network extracted based on our proposed Hellinger distance provides valuable information about trustors' connections and their trust behavior. In this figure, the comparison of similarity measures in combination with both degree and BLC centrality measures shows that the connection similarity has the worst RMSE by increasing $\beta$, and Hellinger similarity has the least RMSE.


Furthermore, according to Equation (\ref{eqn:linfun}) and Figure \ref{fig:all_rmse}, which contains all the different combinations of similarities and centralities, it is shown that by increasing the $\beta$ value, consequently by decreasing the impact of user centralities, the accuracy of prediction protocol strictly increases. It illustrates that considering nodes' centrality might not help the prediction mechanism. Hence, we conclude that although nodes' centrality might not strongly represent their trustworthiness, their similarity reveals their trust pattern similarity effectively. Briefly speaking, it affirms that incorporating nodes' similarities enhances the effectiveness of the prediction mechanism significantly.



\begin{figure*}[]
\vspace{-3mm}
\minipage{0.3\textwidth}
\includegraphics[width=\linewidth]{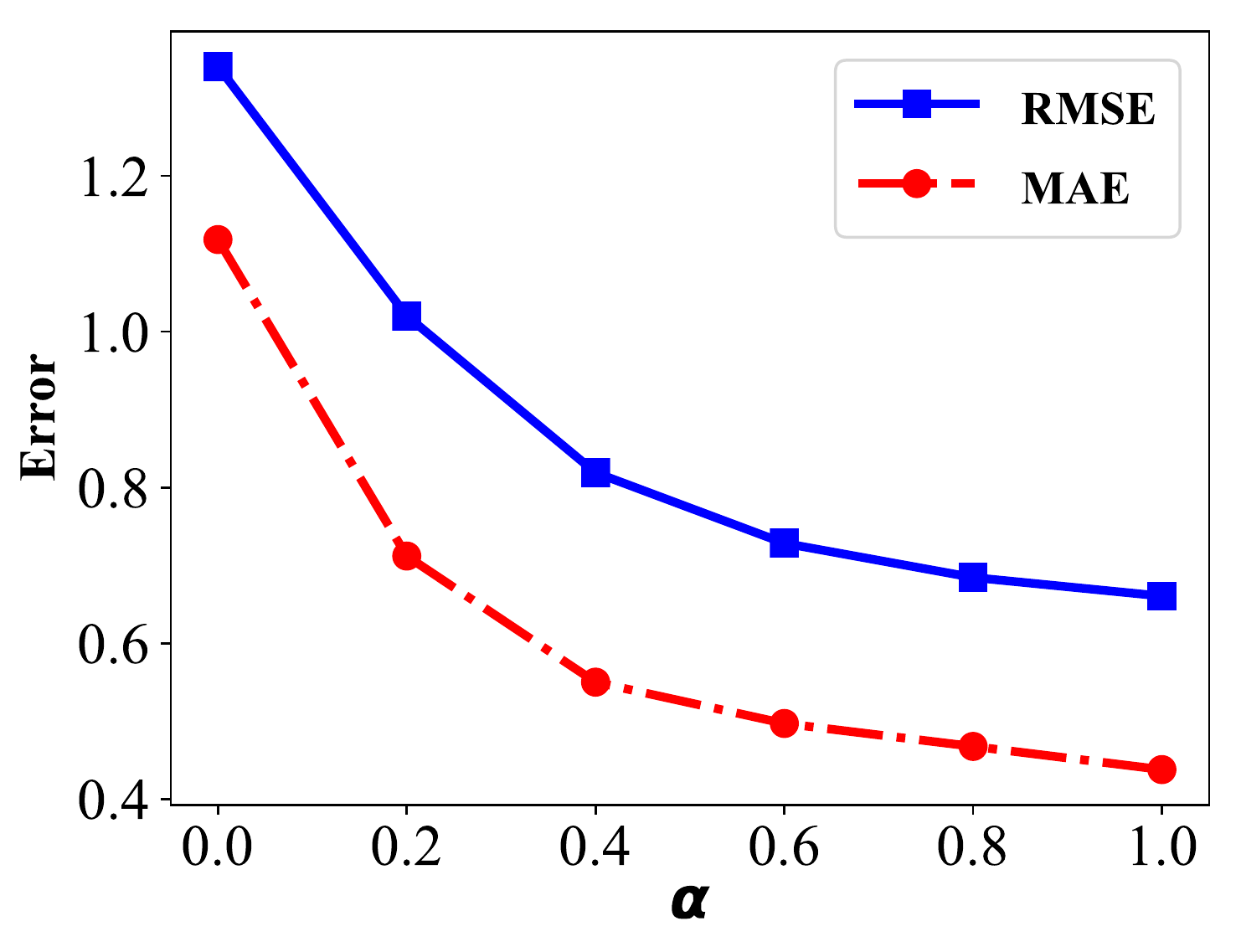}\vspace{-4mm}
\caption{\small Errors of our prediction mechanism using the best setting for different values of $\alpha$.}\label{fig:alpha}
\endminipage\hfill
\minipage{0.3\textwidth}

 \includegraphics[width=\linewidth]{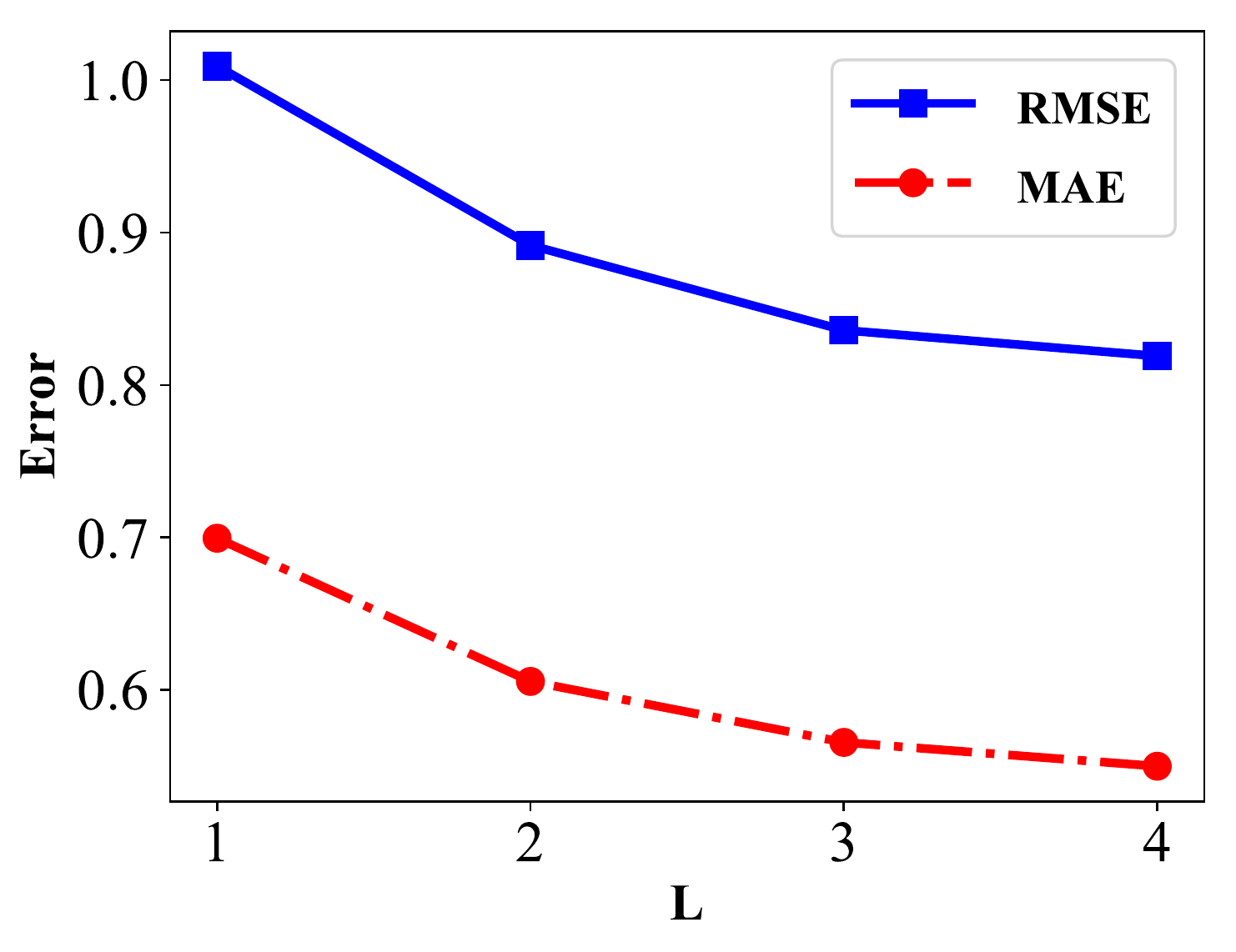}\vspace{-4mm}
 \caption{\small Errors of our prediction mechanism using the best setting for different latent features size $L$.}\label{fig:latent}
\endminipage\hfill
\minipage{0.3\textwidth}

 \includegraphics[width=\linewidth]{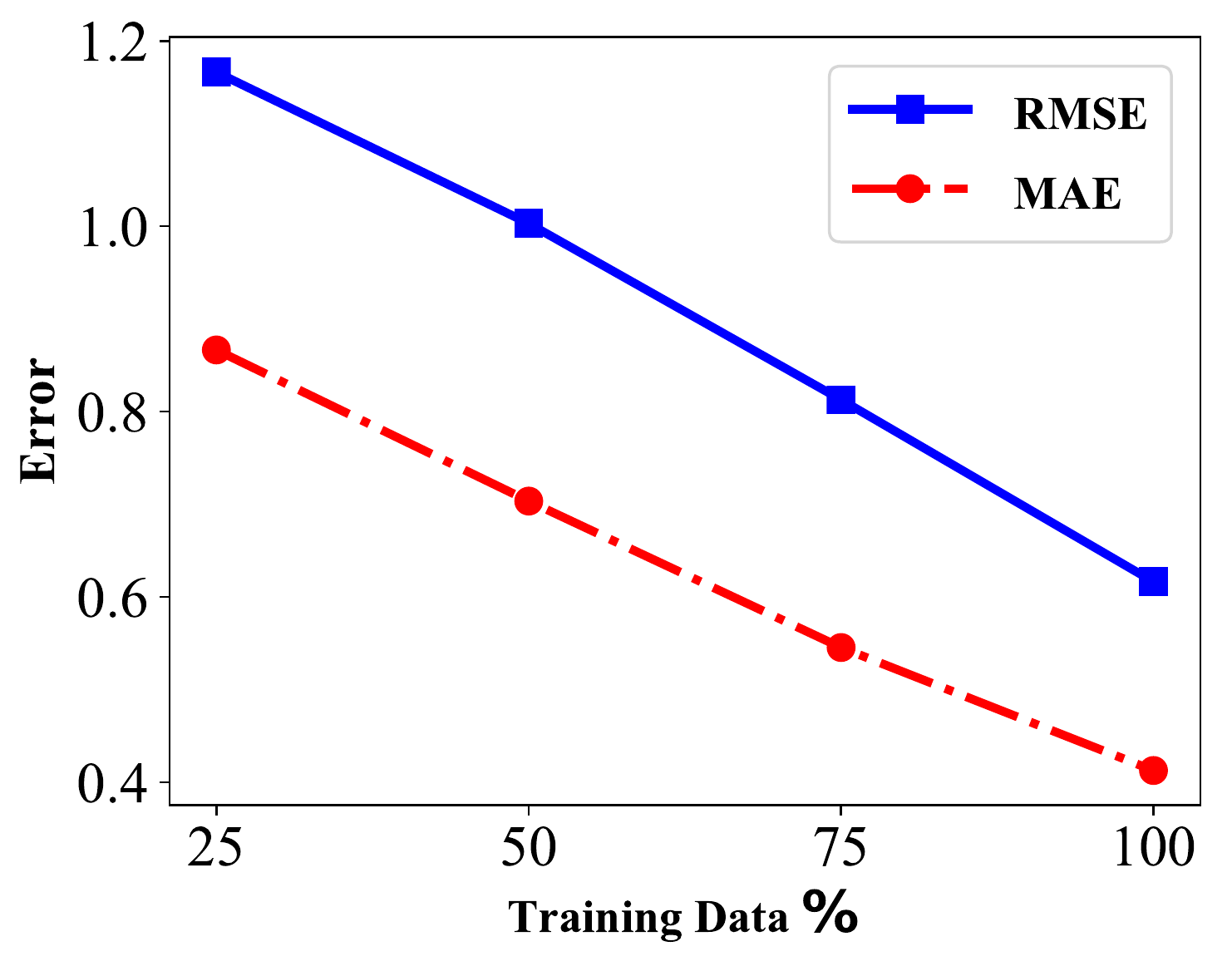}\vspace{-4mm}
 \caption{\small Errors of our prediction mechanism using the best setting for various train versus test splits.}\label{fig:training}
\endminipage
\vspace{-6mm}
\end{figure*}

In order to analyze how much node's friends might affect the prediction protocol, one can leverage the idea of the \textit{Elbow} method. This method is generally designed to discover the best number of clusters in \textit{Clustering Analysis}; here, it helps find the best value of $\alpha$ in Equation (\ref{eqn:rabete_dah}).
In Figure \ref{fig:alpha}, we can see that until $\alpha=0.4$, the gradients of both lines are high, but from there on, adding to $\alpha$ does not help the precision much more. Hence, $\alpha=0.4$ is selected as the best setting.

As depicted in Figure \ref{fig:latent}, increasing latent features' size improves our prediction accuracy; however, it imposes extra overhead because of larger latent feature matrices. Due to our prediction protocol application, choosing the appropriate latent feature size by balancing between having more accuracy or a light-weight process is possible. We use $L=4$ to learn \textit{L}-dimensional (trustors and trustees) latent feature matrices since it seems that the testing error reaches its expected (true) error value at this value ($L=4$), which is not too large to cause excessive overhead.


Figure \ref{fig:training} shows that using more training data generally improves the prediction's precision, which is somehow obvious. In this figure, we used the best setting (Hellinger as similarity, $\beta=1$, $\alpha=0.4$, and $L=4$) in our prediction mechanism. In general, there is not an approved ratio (percentage) for the division of training versus testing sets. Less testing data results in a non-generalized model (\textit{i.e.,} high variance). On the other hand, less training data causes that training loss no longer bears relation to test loss (\textit{i.e.,} high bias) and brings overfitting problems. However, this problem becomes less severe when the size of training data increases. Therefore, between 90:10, 80:20, and 75:25 splits, we chose 75:25 split since our dataset is large enough.

%

\begin{figure}[]
\vspace{-1mm}
\centering
\captionsetup{justification=centering}
\includegraphics[width=0.48\textwidth]{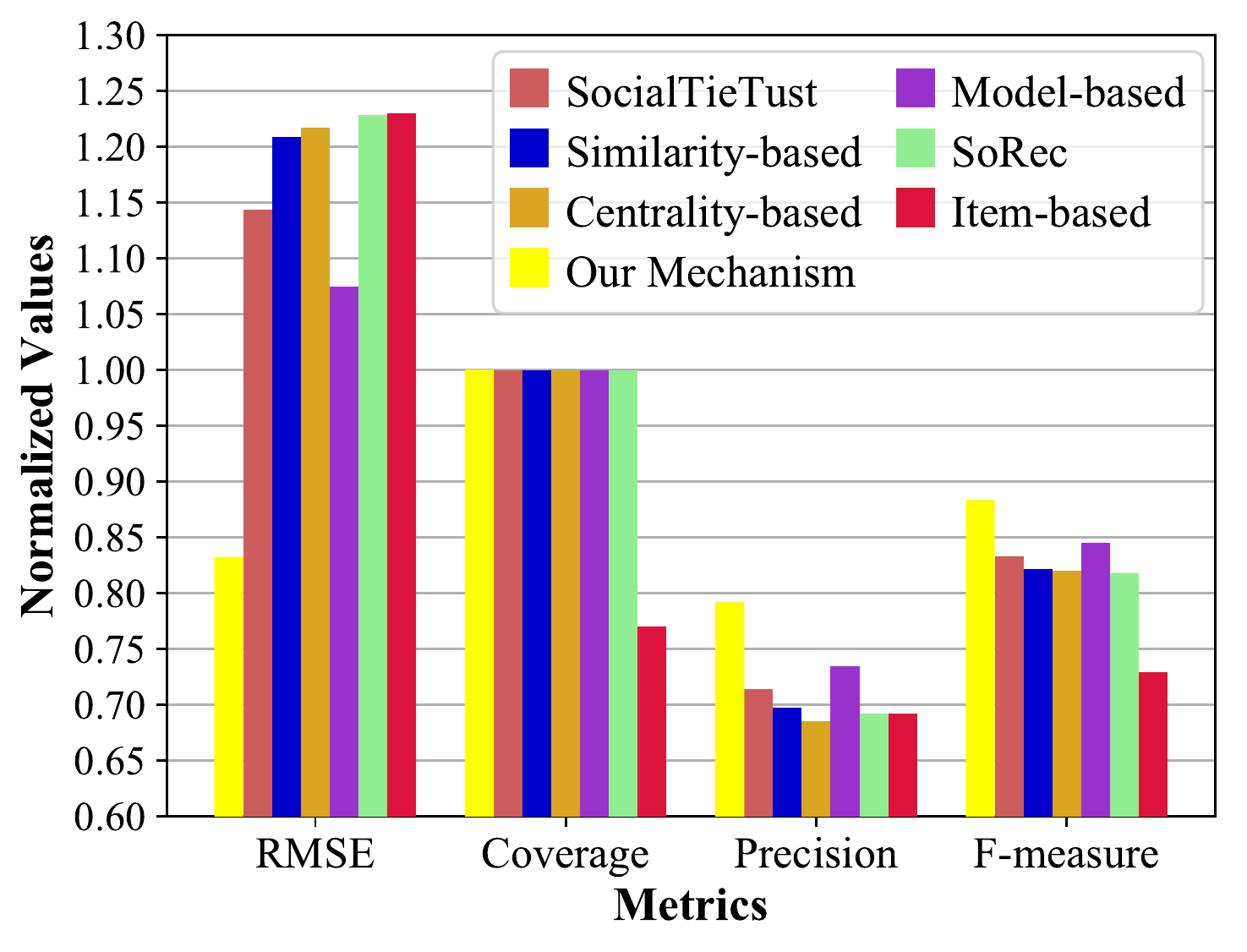}
\vspace{-2mm}
\caption{\small Comparison of our method using its best setting ($L=4$, $\alpha=0.4$, $\beta=1$, and Hellinger as similarity) with other methods.}
\label{fig:comp}
\end{figure}



To show the effectiveness of our prediction method in comparison with the best existing methods, we evaluate the methods in terms of RMSE, coverage, precision, and F-measure. To the extent of our knowledge, there is no other IoT/SIoT trust management model that utilizes a matrix factorization mechanism. Therefore, we compared our method with relevant social network methods. The best existing methods in the literature for comparison are (1) SocialTieTrust \cite{davoudi2018social}, (2) Similarity-based \cite{davoudi2016modeling}, (3) Centrality-based \cite{davoudi2016modeling}, (4) Model-based \cite{jamali2010matrix}, (5) SoRec \cite{ma2008sorec}, and (6) Item-based \cite{sarwar2001item}.

As shown in Figure \ref{fig:comp}, our method clearly provides very low RMSE, compared to the competing methods with a considerably higher RMSE. As Koren states in \cite{koren2010factor}, even a small improvement in RMSE could have significant enhancements. Since for all competing methods, $RMSE_{max}$ is equal to 4, and according to the combination of RMSE and $RMSE_{max}$, our method is selected as the best candidate in terms of precision. Model-based and SocialTieTrust are the second and third best methods, respectively, with slight differences (in all terms). Additionally, our proposed method outperforms all the other competing methods in terms of F-measure. Eventually, our proposed method's coverage is $100\%$ that is the best coverage a prediction mechanism can provide. Other methods, except for Item-based, provide $100\%$ coverage too. The aforementioned results appear to be better than all the previously obtained results. Therefore, we conclude that the implicit social information from Hellinger distance can be incorporated into the matrix factorization mechanism to make effective predictions.

Suppose the number of trustors is $n$, and the highest degree of trustees in the bipartite graph is $\Delta$. Considering that the bi-adjacency matrix $B$ of the SIoT model and adjacency matrix $A$ of trustors' social network are sparse matrices, the most time-consuming part of our algorithm takes $O(n^2 \Delta)$, which is related to generating the Hellinger distance matrix. Our mechanism is more efficient than Similarity-based \cite{davoudi2016modeling}, Centrality-based \cite{davoudi2016modeling}, and SocialTieTrust \cite{davoudi2018social} with time complexity $O(n^3)$. Also, our algorithm may not be better than Model-based \cite{jamali2010matrix}, Item-based \cite{sarwar2001item}, and SoRec \cite{ma2008sorec} that respectively have the complexity of $O(n d^2)$, $O(m n)$, and $O((z+r) \times I)$, where $d$ is the average degree in the social network of trustors, $m$ is the number of items, $z$ and $r$ are the numbers of non-zero entries of rating matrix and social network adjacency matrix, and $I$ is the number of iterations for convergence. Nonetheless, not only is our mechanism not considerably time-consuming compared to the aforementioned models, but also its accuracy is way better than the competing methods.
\vspace{-2mm}
\subsection{Trust Protocol Performance}
\label{trustprotocolperformance}
As we mentioned in Section \ref{relatedwork}, there are few trust management methods with which we can compare our method. Hence, we follow the same simulation evaluation strategy as in \cite{chen2016trust}, which is one of the most outstanding works in the SIoT trust management literature. Although we follow the same evaluation approach as \cite{chen2016trust}, our methods are not fully comparable because the described attacks are not quite identical; moreover, the architectures are different, as explained earlier in Section \ref{relatedwork}. Our goal is to investigate our trust management mechanism's performance using the best setting, as analyzed in the previous subsection in a hostile SIoT environment. 
\vspace{-2mm}

\begin{figure*}[ht!]
\vspace{-6mm}
\captionsetup{justification=centering}
\subfloat[Toward a randomly picked benign trustee]{\includegraphics[width=0.32\textwidth]{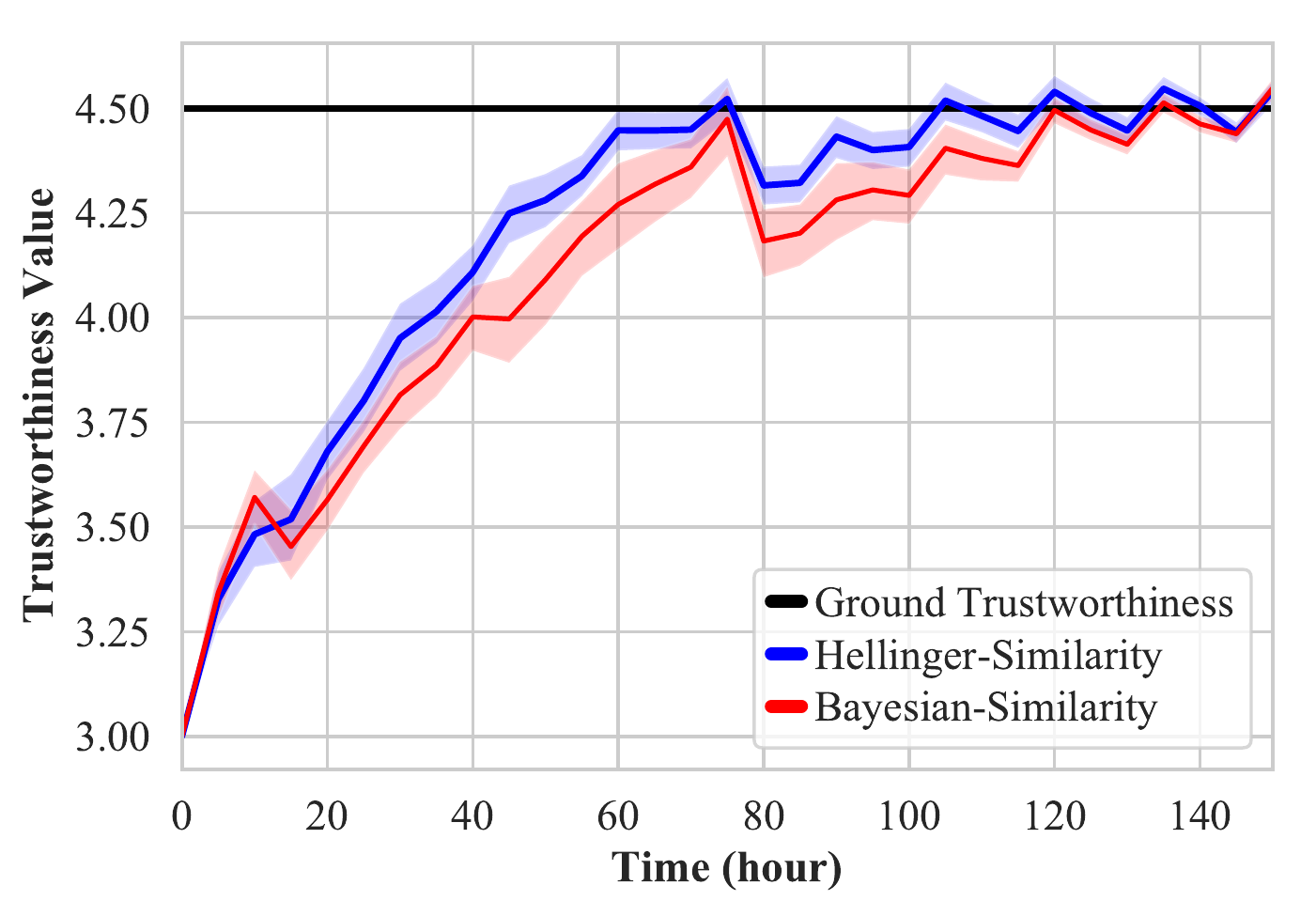} \label{fig:firstgood}}
\hspace{\fill}
\subfloat[Toward a randomly picked malicious trustee (not performing opportunistic service attack)]{\includegraphics[width=0.32\textwidth]{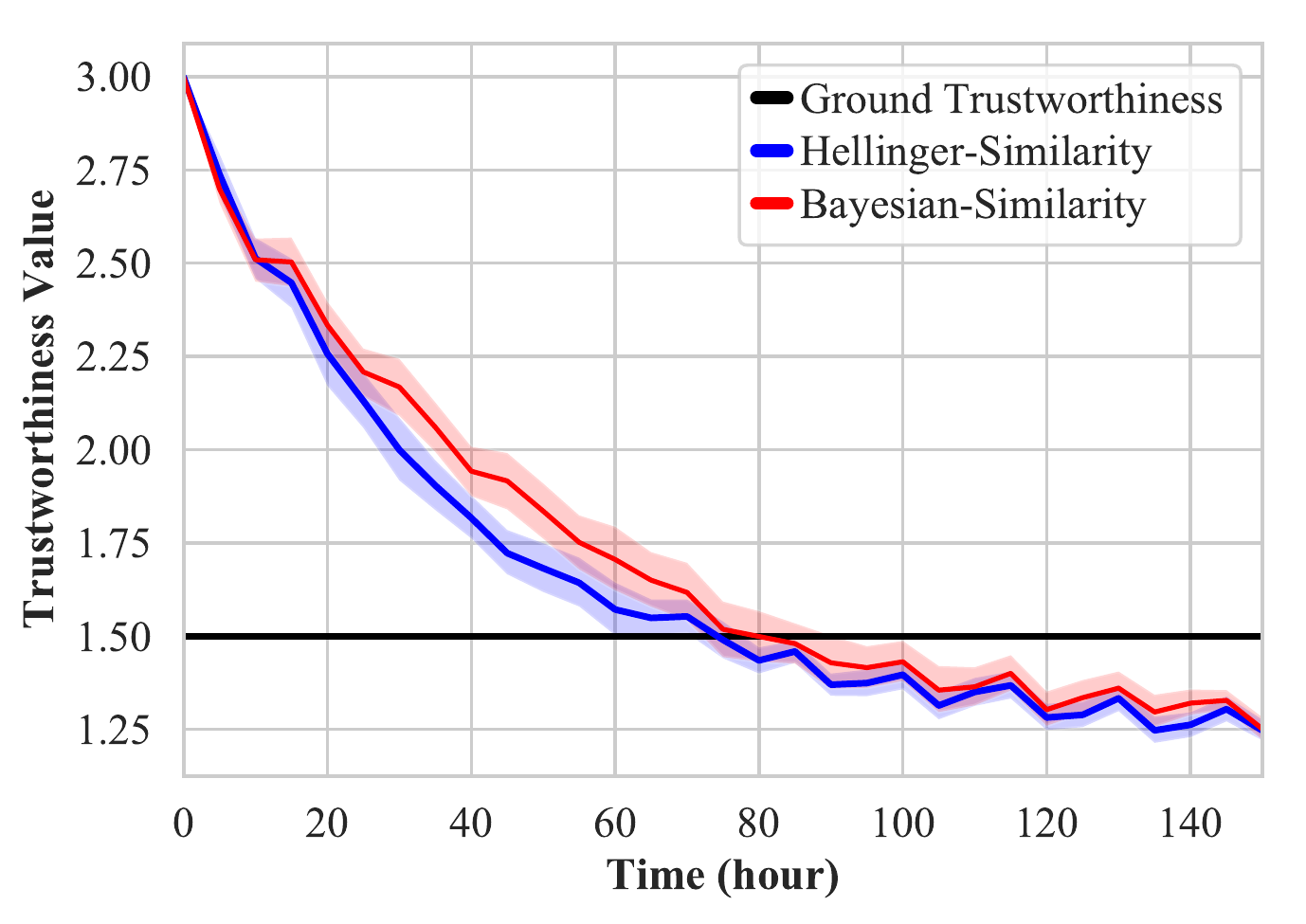} \label{fig:firstbad}}
\hspace{\fill}
\subfloat[Toward a randomly picked malicious trustee (performing opportunistic service attack)]{\includegraphics[width=0.32\textwidth]{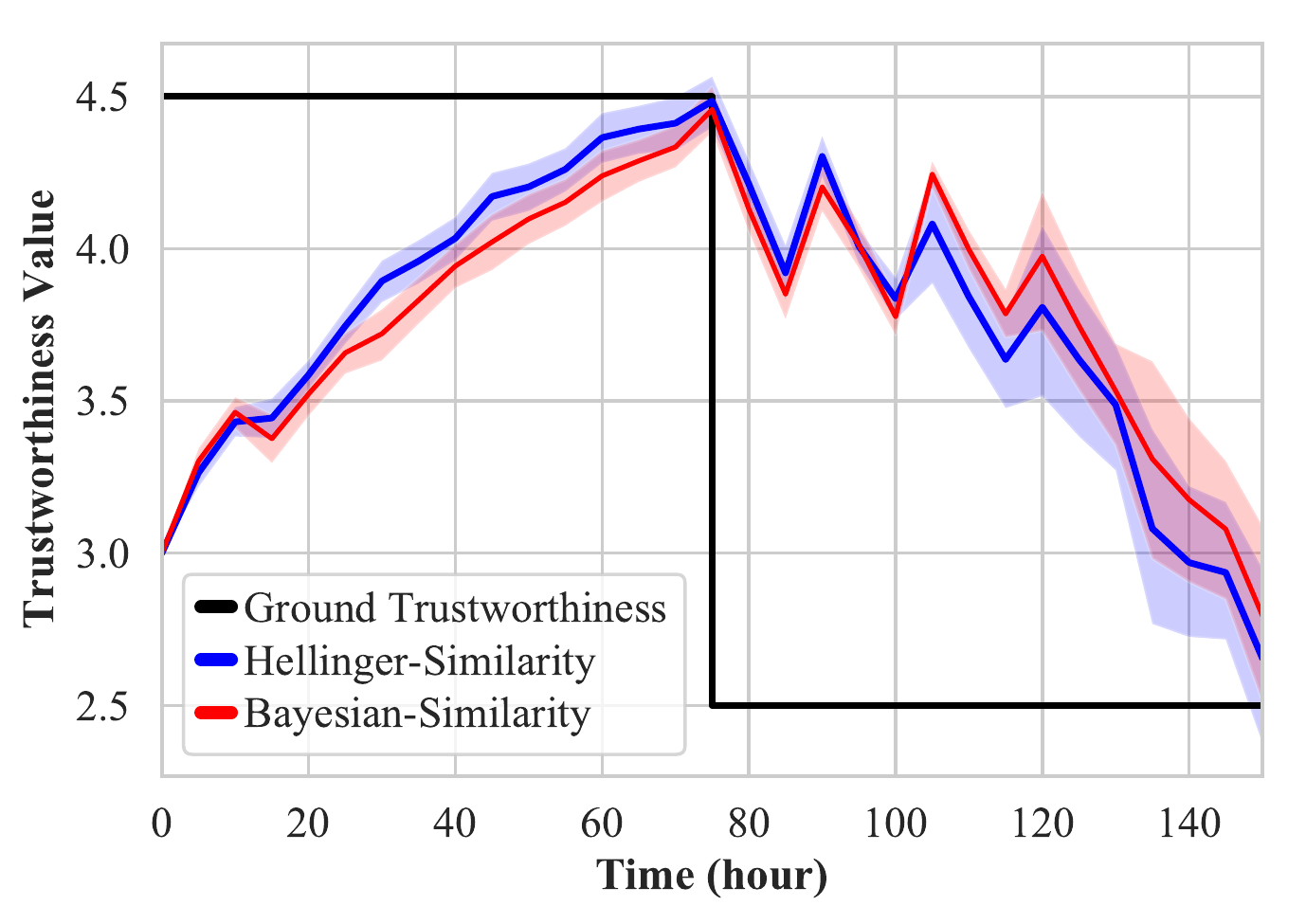} \label{fig:firstugly}}\\ \vspace{-3mm}
\caption{\small Effect of similarity measures on the trustworthiness evaluation using the best setting.} \label{fig:malkhoobbadzesht111}
\vspace{-9pt}
\end{figure*}

\subsubsection{Settings \& Metrics}
\label{settingsmetrics1}
We conducted several experiments through $150$ hours of simulation to validate the convergence, accuracy, and attack resiliency properties. To determine the interaction-contact time, our simulation's interaction pattern follows the setting described in \cite{karagiannis2010power}, which is close to real traces.

We consider a SIoT environment with maliciousness factor ${\lambda\in[10\%, 50\%]}$ that will be selected randomly in each simulation execution. The maliciousness factor $\lambda$ indicates the ratio of malicious nodes to all nodes. A malicious node can perform any kind of attack, addressed in section \ref{problemstatement}. Our SIoT simulation environment consists of $70$ trustees and $100$ trustors, which form $14$ and $20$ physical groups, respectively. Instances of these physical groups could be smart homes or smart vehicles that have one non-lightweight central node and four lightweight SIoT sensors.

During the simulation, the trustor groups form a social network of trustors, as explained in Section \ref{socialnetworkoftrustors}. They then perform the matrix factorization mechanism and predict missing values of the bi-adjacency matrix $B$, as detailed in Section \ref{predictionmechanism}, every $24$ hours. We believe that performing matrix factorization once a day (\text{i.e.,} every 24 hours) is not computationally a heavy task.
Each SIoT node has an objective (ground) trustworthiness value in the interval $(0, 1]$ \cite{chen2016trust} which would be specified randomly with respect to the level of its malicious behavior at the beginning of the simulation. However, we scale this interval to $[1, 5]$ in order to be coherent with the last evaluation subsection and have more readable figures.
For example, malicious nodes receive objective trustworthiness values closer to the minimum of the interval (\text{i.e.,} $1$), and non-malicious nodes acquire objective trustworthiness values near the maximum of the interval (\text{i.e.,} $5$). The trustworthiness value of SIoT nodes remains unchanged during the simulation except for malicious nodes that perform opportunistic service attack.
As mentioned earlier in Section \ref{problemstatement}, trustors measure and rate the trustworthiness of trustees based on their service usage experiences. This rating scales between $1$ and $5$ and would be maintained to benefit future decisions. Our proposed trust management mechanism sets all SIoT nodes' initial trustworthiness to the middle of its range (\text{i.e.,} $3$).
\vspace{-2mm}

\subsubsection{Evaluation Results}
\label{results1}
In this subsection, the two best similarity methods, Hellinger and Bayesian, will be studied first. Then, we investigate the effect of hostility changes on the trustworthiness evaluation operation. In this experiment, we analyze our selected mechanism's performance through trust evaluation results of trustor nodes toward three trustee nodes randomly picked from a $30\%$ hostile SIoT environment. The first trustee node is a benign node whose objective trustworthiness is equal to $4.5$ (out of $[1, 5]$). The second and third trustees are malicious nodes. The former malicious node does not perform an opportunistic service attack, and its trustworthiness value remains fixed on $1.5$. Nevertheless, the latter performs opportunistic service attack, and its objective trustworthiness value decreases from $4.5$ to $2.5$ in the middle of the simulation.

%

Figure \ref{fig:firstgood} shows trustworthiness evaluation toward the non-malicious node. The pale area around lines exhibits the empirical confidence intervals with $90\%$ confidence. The trustworthiness value of trustors toward this trustee node starts at $3$ and approximates to $4.5$, which is the benign trustee's trustworthiness value. Predictably, the Hellinger-Similarity setting converges with more confidence and faster than the Bayesian-Similarity setting, as expected from Figure \ref{fig:all_rmse}. Furthermore, we observe that as trustworthiness converges, it fluctuates around the objective trustworthiness with more confidence.


Figure \ref{fig:firstbad} demonstrates trustworthiness evaluation toward a random malicious node that does not perform an opportunistic service attack. As it should, the trustworthiness value decreases to approach the objective value. Again, we can observe that the Hellinger-Similarity setting performs better than the Bayesian-Similarity setting. However, the difference between the two similarity settings is not significant. The other phenomenon that attracts our attention is that the mechanism underestimates the trustworthiness value after reaching objective trustworthiness. The reason for this phenomenon is that trustors do not trust and use malicious trustees just after the trustworthiness value decreases; hence, no further usage experience affects the trustworthiness value. However, the matrix factorization mechanism still reduces the trustworthiness value due to the trustors' past experiences.

Figure \ref{fig:firstugly} demonstrates trustworthiness evaluation toward an opportunistic service attacker node. Convergence rates of different settings are as we expected from the previous part. Although the convergence speed and confidence became lower after the malicious node changed its behavior, we can see that our protocol can proficiently track the malicious trustee's trustworthiness value. The reason for high fluctuation after $t=75$ is that the system gets confused about the behavior change, but after $45$ hours, the procedure monotonically decreases.

\begin{figure*}[]
\vspace{-5mm}
\captionsetup{justification=centering}
\subfloat[Toward a randomly picked benign trustee]{\includegraphics[width=0.32\textwidth]{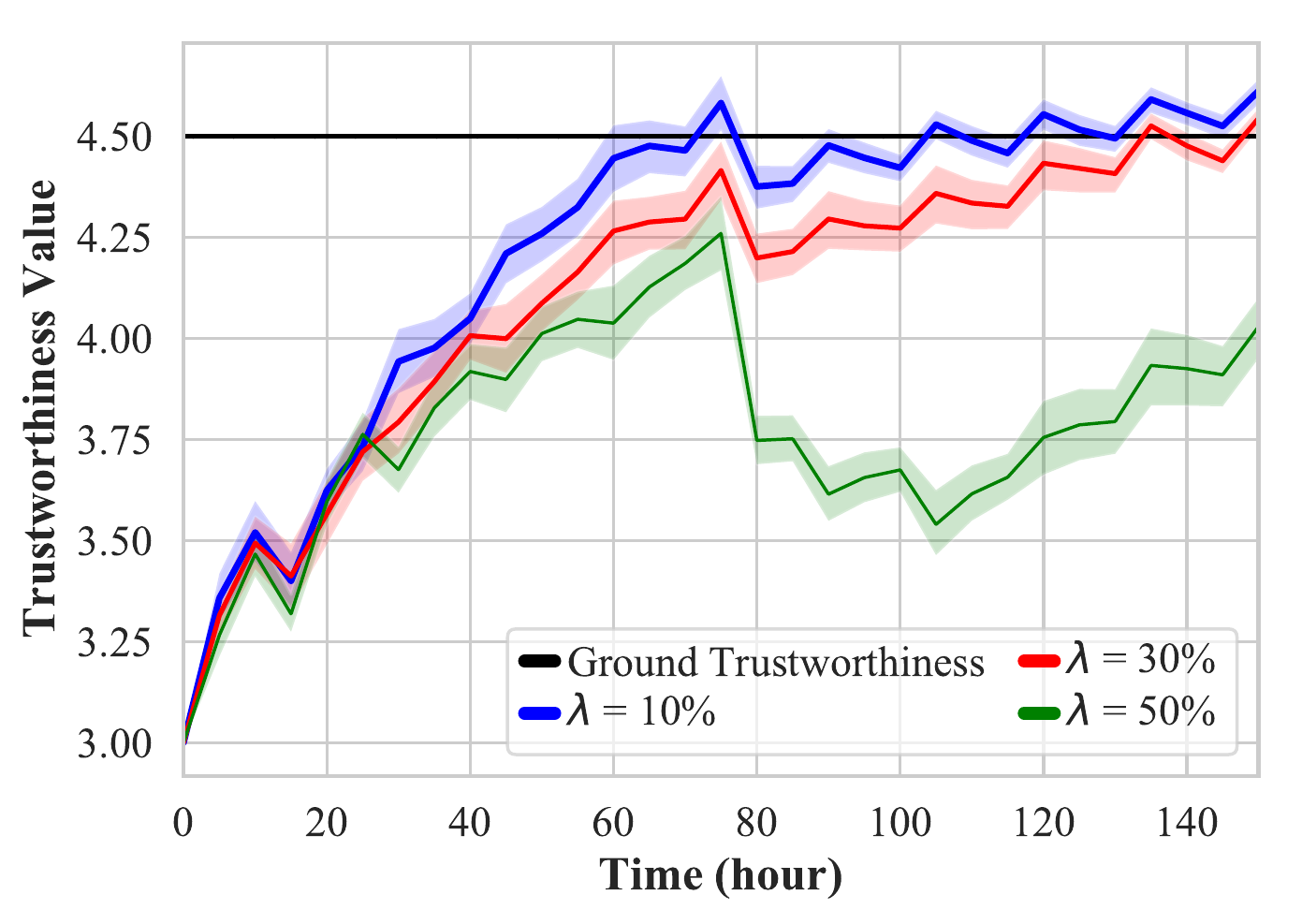} \label{fig:malkhoob}}
\hspace{\fill}
\subfloat[Toward a randomly picked malicious trustee (not performing opportunistic service attack)]{\includegraphics[width=0.32\textwidth]{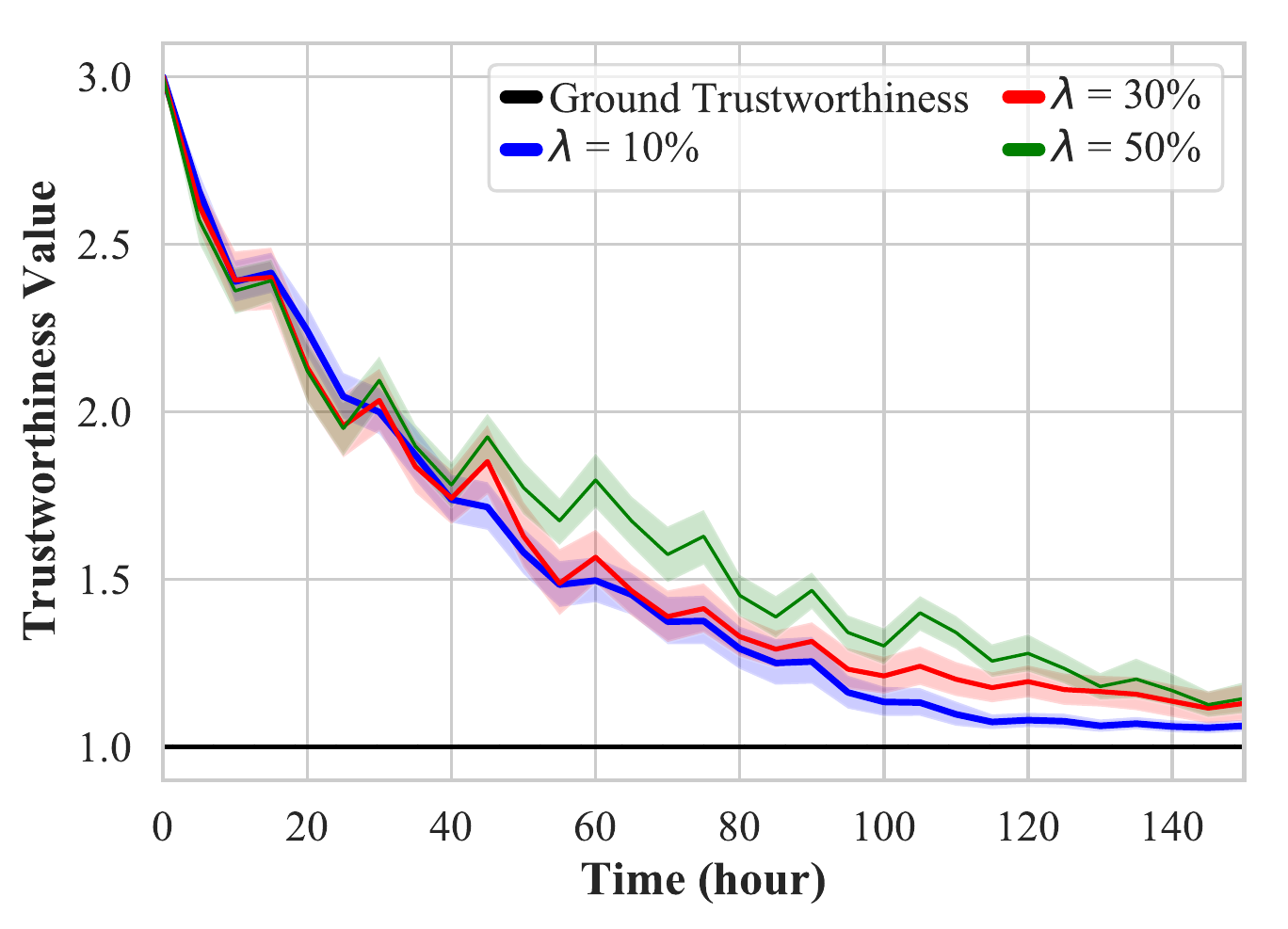} \label{fig:malbad}}
\hspace{\fill}
\subfloat[Toward a randomly picked malicious trustee (performing opportunistic service attack)]{\includegraphics[width=0.32\textwidth]{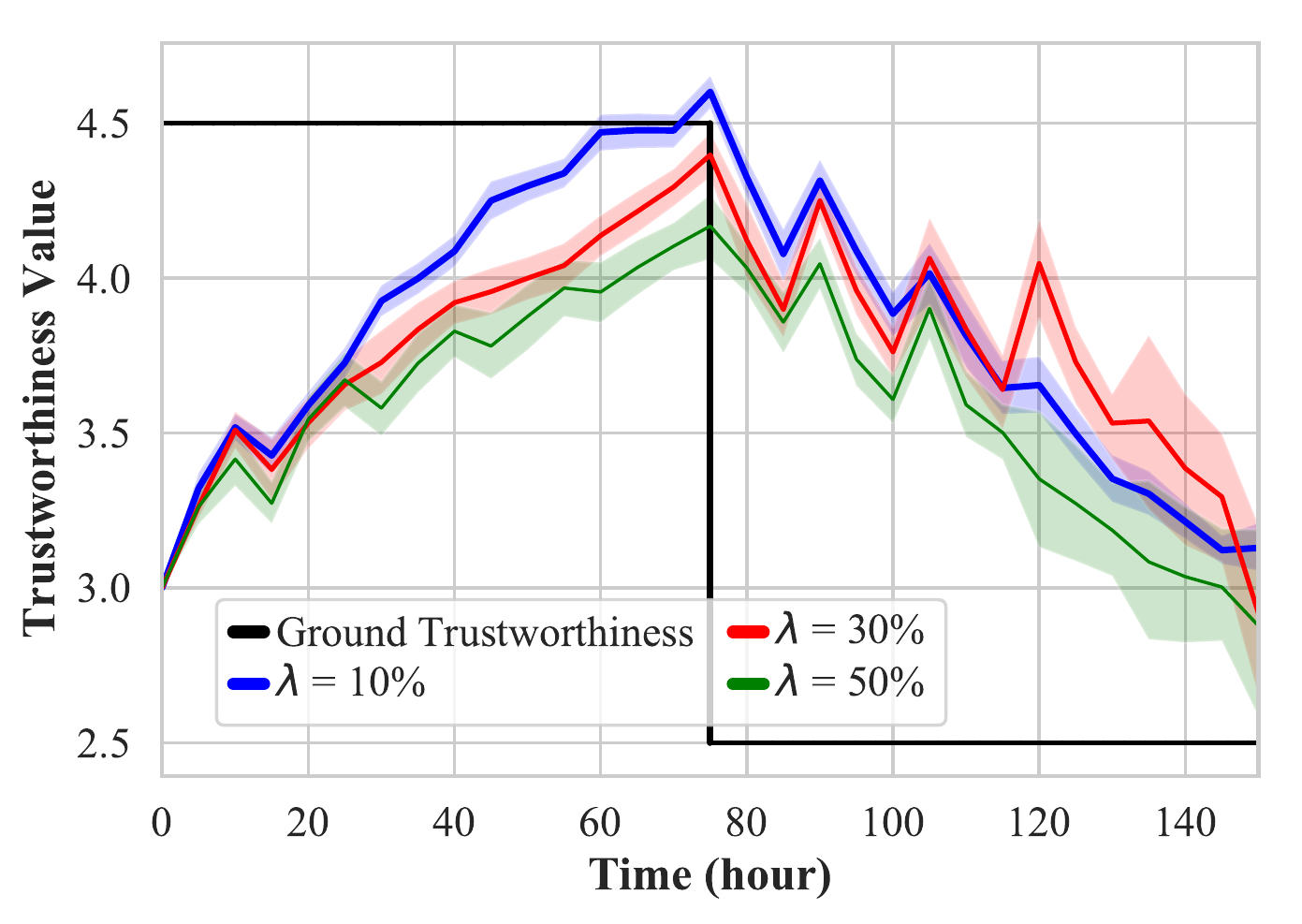} \label{fig:malzesht}}\\ \vspace{-3mm}
\caption{\small Effect of malicious factor $\lambda$ on our trust management efficiency using the best setting.} \label{fig:malkhoobbadzesht}
\vspace{-5mm}
\end{figure*}

%
%
%
 
Next, we investigate the effect of malicious factor $\lambda$ on our mechanism efficiency. For sensitivity analysis, we change the SIoT environment hostility from low hostile ($\lambda=10\%$) to a very hostile environment ($\lambda=50\%$). Figures \ref{fig:malkhoob} to \ref{fig:malzesht} demonstrate the trustworthiness value of trustors toward randomly picked trustees through our best trust management mechanism setting, which utilizes Hellinger as the similarity and no effect of centrality ($\beta=1$).

Figure \ref{fig:malkhoob} shows trustworthiness evaluation toward a benign trustee. We can see that all three lines approach the (trustee's) objective trustworthiness value. However, we observe that the green line drops excessively at $t=75$ because of lots of malicious nodes that perform opportunistic service attack and change their behavior at that time. However, it still converges on the objective trustworthiness value, as we expected. An interesting observation in Figure \ref{fig:malkhoob} is that for $\lambda=10\%$, our proposed trust management overestimated the trustworthiness value of a trustee; however, it preserves the order of trustees in terms of their trustworthiness values.

Moreover, Figure \ref{fig:malkhoobbadzesht} indicates that our trustworthiness evaluation mechanism converges toward a trustee's objective trustworthiness accurately yet quick. We can see that our mechanism confidence and convergence rate are higher in a less hostile environment. However, as the malicious factor increases, the protocol performance is still acceptable. These results demonstrate our mechanism's high resiliency toward various attacks, even in a highly hostile SIoT environment.

\vspace{-2mm}
\subsection{SIoT Real Application Performance}
\label{siotapplicationperformance}
This section evaluates the effectiveness of our proposed trust management mechanism through a real-world SIoT use case \cite{ni2018providing}. We aim to run our protocol on top of such an end-to-end scenario to validate its robustness against the cold start problem and related attacks. We compare our system's performance with a random system in which trustors select their service providers among all available trustees randomly.

\vspace{-3mm}
\subsubsection{Settings \& Metrics}
\label{settingsmetrics2}
We consider a smart city in which people benefit from a smart health-care system. In this city, individuals have health-care applications installed on their mobile phones. One of the purposes of this health-care system is to provide people with air pollution information. Suppose that Alice, who is a respiratory patient, wants to go jogging. Alice's doctor advises her not to get into polluted areas. In this case, she lets her smartphone connect to sensor devices in an area she is about to step into and alert her if any air pollution is detected. We want to analyze how her smartphone behaves in this situation.

\begin{figure}[]
 \vspace{-2mm}
 \centering
 \captionsetup{justification=centering}
 \includegraphics[width=0.49\textwidth]{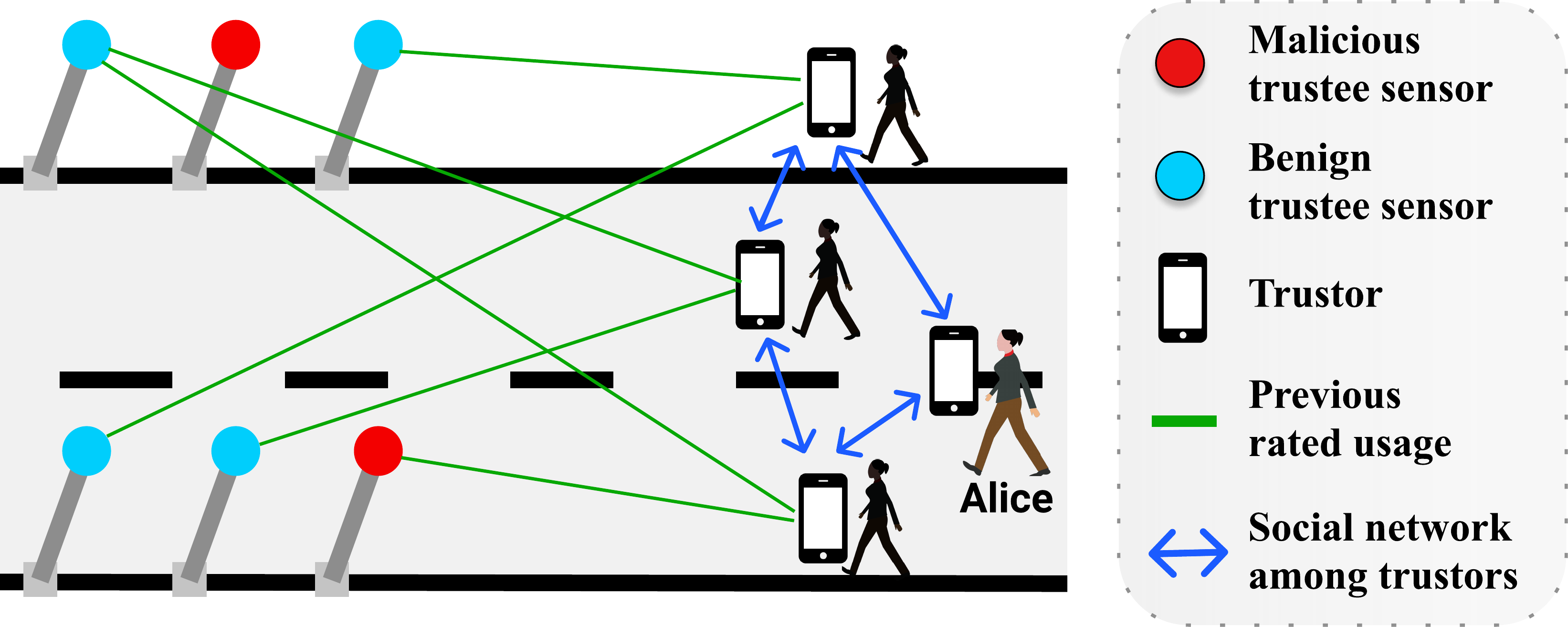}\vspace{-2mm}
 \caption{\small Illustration of the simulated use-case}\label{fig:alice}
 \vspace{-1mm}
 \end{figure}

Her smartphone, as a trustor, is entirely strange in the environment that she runs in. Besides, many malicious and imperfect SIoT sensors provide wrong or inaccurate data; for instance, in this simulation, there are malicious nodes that perform the ballot stuffing attack to boost the trustworthiness value of a trustee node with objective trustworthiness equal to $2$.
Thus, the trust management mechanism that her smartphone runs to decide which of the new trustees are reliable must \textbf{overcome cold start problem} and be \textbf{resilient to various attacks}. Figure \ref{fig:alice} depicts the scenario mentioned above. 

\subsubsection{Evaluation Results}
\label{results2}
\begin{figure}[]
 \vspace{-6mm}
 \centering
 \captionsetup{justification=centering}
 \includegraphics[width=0.46\textwidth]{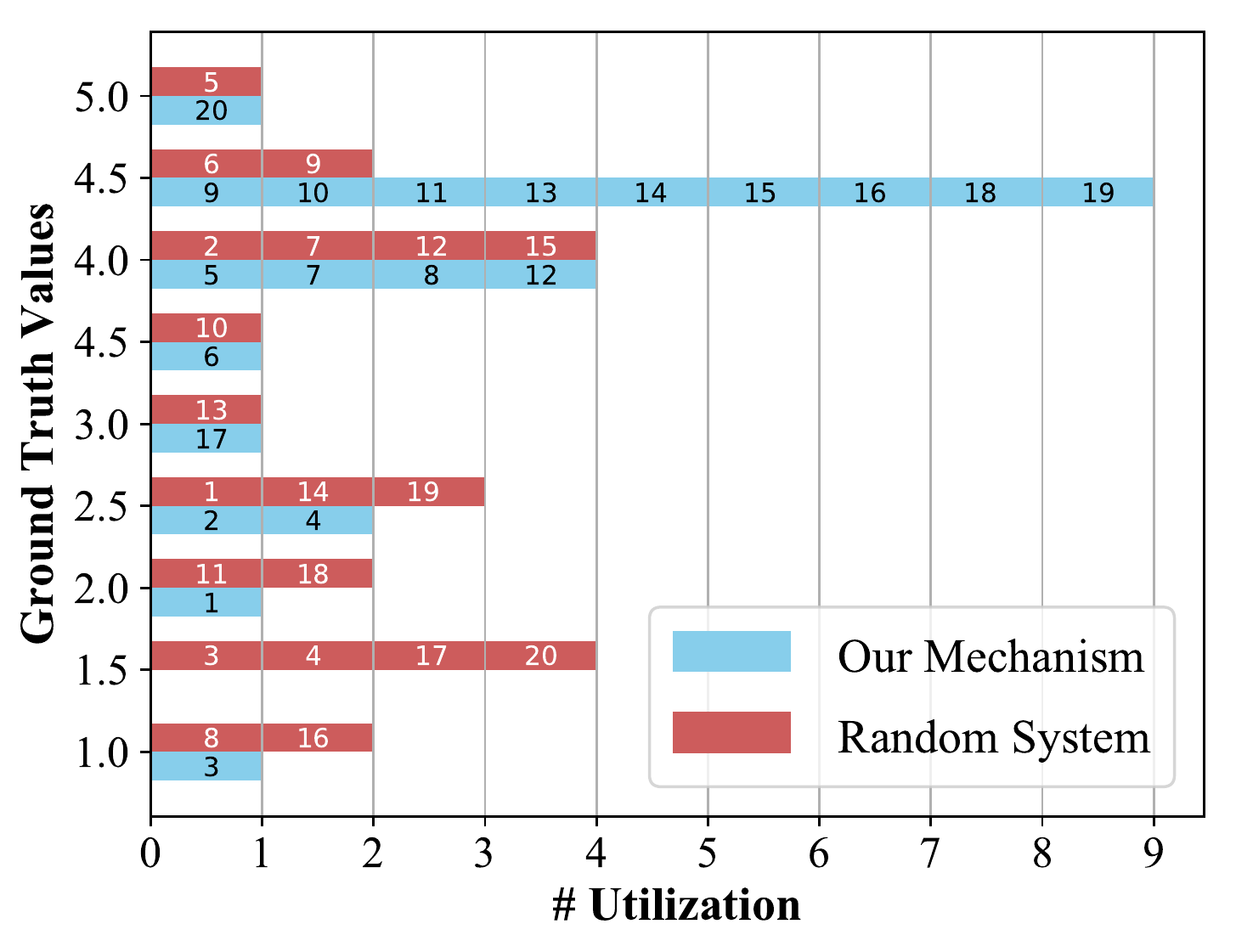}\vspace{-5mm}
 \caption{\small Performance comparison of our proposed trust management versus a random system in a real-world SIoT application scenario. Y-axis shows ground trustworthiness values that represent trustee groups, and X-axis shows the number of times that each system selects each group. The numbers on bars show the order of selecting each group.}\label{fig:OOO}
 \vspace{-2mm}
 \end{figure}

Alice's smartphone first builds a social network among other trustors to discover trustors' trust pattern similarities in this simulation. It then utilizes the described matrix factorization technique to find the most trustworthy sensor (trustee). Figure \ref{fig:OOO} compares our mechanism against a random method that randomly picks trustees without considering the (trustees') trustworthiness values. We classify trustees by their objective trustworthiness values into nine groups. Figure \ref{fig:OOO} shows how many times Alice's smartphone has used each group. We can see that at the beginning of the simulation, the trustor experiments with various trustees to learn about their behaviors; then, it keeps using the most trustworthy trustees, which has objective trustworthiness equals to $4$ and $4.5$. Also, it is represented that the group with ground trustworthiness equals $5$ has been utilized just once out of $20$. The reason is that there is just one trustee in such a perfect group, and Alice's smartphone did not discover it until the last minutes of the simulation.

Furthermore, we can see that our method is resilient to malicious nodes performing the ballot stuffing attack since our system selected malicious trustee with objective trustworthiness equal to $2$ only once. To explain more, although other trustors promote that malicious trustee node by faking good experiences, they have no effect on Alice's smartphone since our system enables the trustor to use other trustors' experiences whose trust patterns are similar to the trustor (i.e., Alice's smartphone).
One can conclude that our mechanism successfully helps trustors detect and exploit trustees with high trustworthiness values, and it definitely outperforms the random model even for newcomer nodes that cause the cold start problem. 

\vspace{-3mm} 
\section{Conclusion}
In this paper, we proposed a novel trust management mechanism in SIoT. We employed Hellinger distance to build a social network of trustors. The social relations in the network show behavioral trust similarity among network nodes. The trustworthiness value of trustees is predicted using both trustor's experience and its friends' feedbacks. In order to utilize the feedbacks, we designed a social trust model using centrality and similarity measures. To the best of our knowledge, it is the first paper using matrix factorization to predict the trustworthiness values of trustees in SIoT.

Our proposed mechanism is globally distributed and considers the data sparsity problems and resource-constraint of IoT devices. We demonstrated the effectiveness of our prediction mechanism by evaluating its accuracy using different settings. We found that the best accuracy occurs when we use inversion of Hellinger distance as the similarity measure in our proposed social trust model, without considering the impact of centrality. Then we compared our prediction mechanism with the best existing methods in the social network literature, and the results showed the superiority of our proposed method in terms of RMSE, coverage, precision, and F-measure. To investigate our trust management mechanism's applicability, we evaluated the convergence, accuracy, and attack resiliency properties of our mechanism's different settings in a hostile SIoT environment. Our simulation results demonstrated that our proposed mechanism accurately converges to the trustee's ground trustworthiness value and resists the malicious nodes. We further showed the utility of our proposed trust management mechanism through a real-world SIoT use case. The simulation showed that our proposed mechanism successfully mitigated the cold start problems, provided resiliency against related attacks, and helped trustors find trustworthy trustees.

As future work, we plan to model the SIoT with hypergraphs because we have found that by representing social relations with naive edges (in traditional graphs), we may lose some information \cite{zhou2007learning, zhang2018beyond}. From the experience of this paper, we believe that finding more meaningful and deeper social relations between SIoT nodes helps us understand their trust pattern similarities better.

\vspace{-1mm}
\bibliographystyle{IEEEtran}
\bibliography{IEEEabrv,references}
\vspace{-45pt}
\begin{IEEEbiography}[{\includegraphics[width=1in,height=1.1in,clip,keepaspectratio]{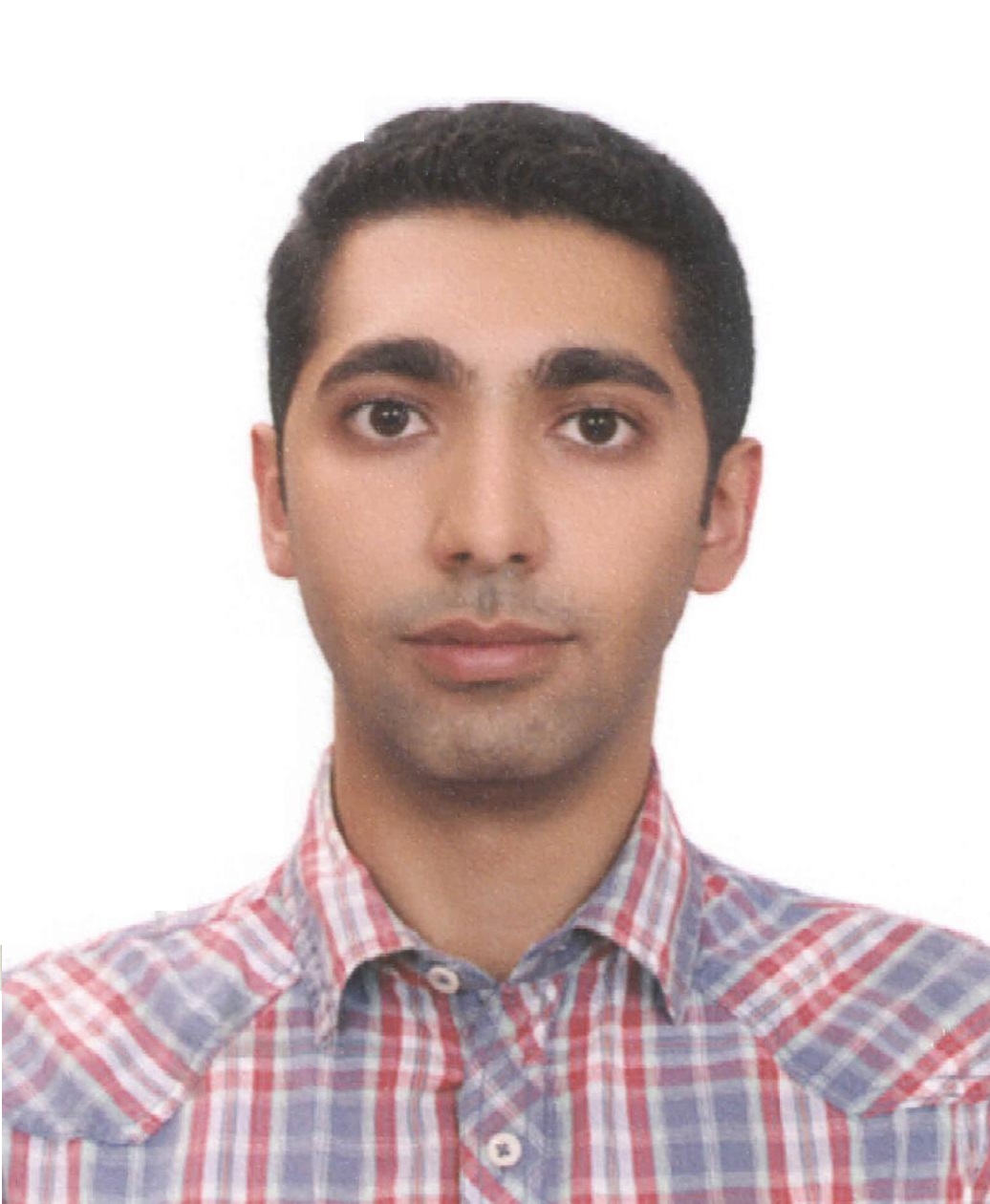}}]{Soroush Aalibagi}
received his B.S. degree in Computer Engineering from Imam Khomeini International University, in 2017 and his M.S. degree from Sharif University of Technology, in 2019. He is currently a research assistant in the Performance and Dependability Laboratory at the Sharif University of Technology. His research interests are in the areas of Social Network Analysis, Graph Theory, Internet of Things, Machine Learning, and Security.
\end{IEEEbiography}
\vspace{-45pt}
\begin{IEEEbiography}[{\includegraphics[width=1in,height=1.1in,clip,keepaspectratio]{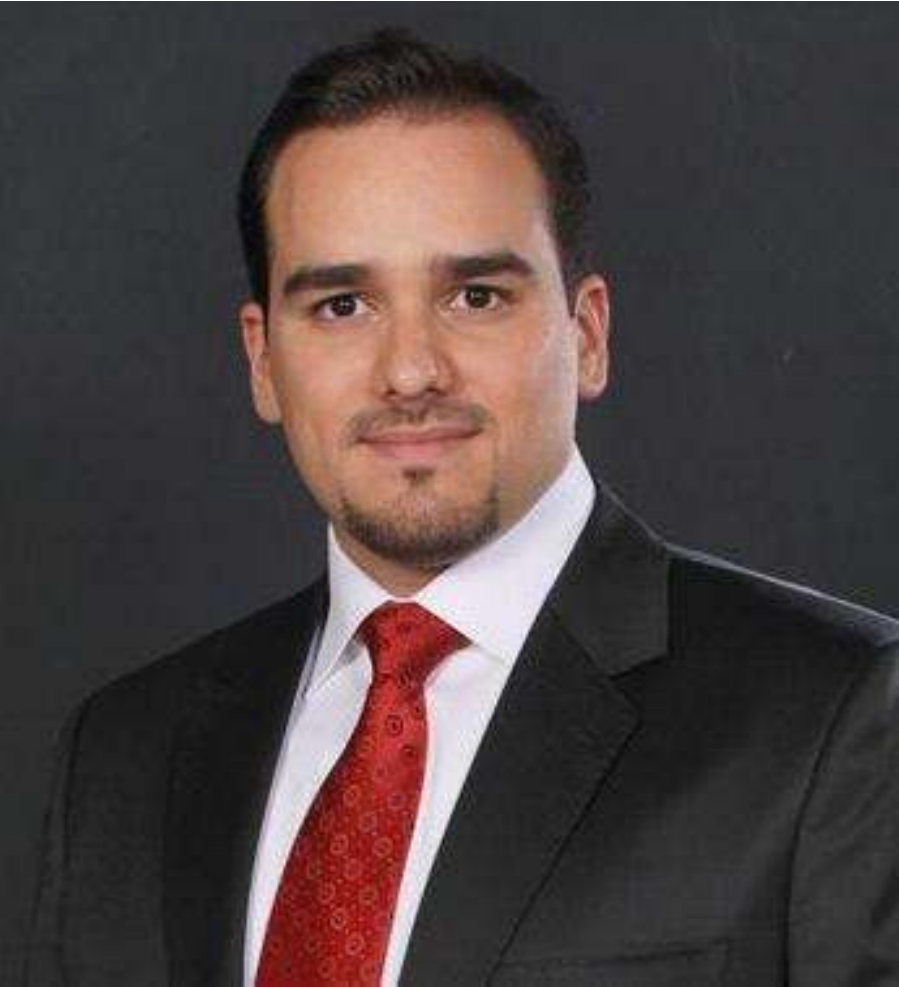}}]{Hamidreza Mahyar}
received his Ph.D. degree in Computer Engineering at the Sharif University of Technology in 2018. He was Postdoctoral Research Fellow at Vienna University of Technology, Austria, and Boston University, USA. His research interests lie in the area of Machine Learning, Deep Neural Networks, Social Network Analysis and Mining, Recommender Systems, and Internet of Things.
\end{IEEEbiography}
\vspace{-45pt}
\begin{IEEEbiography}[{\includegraphics[width=1in,height=1.1in,clip,keepaspectratio]{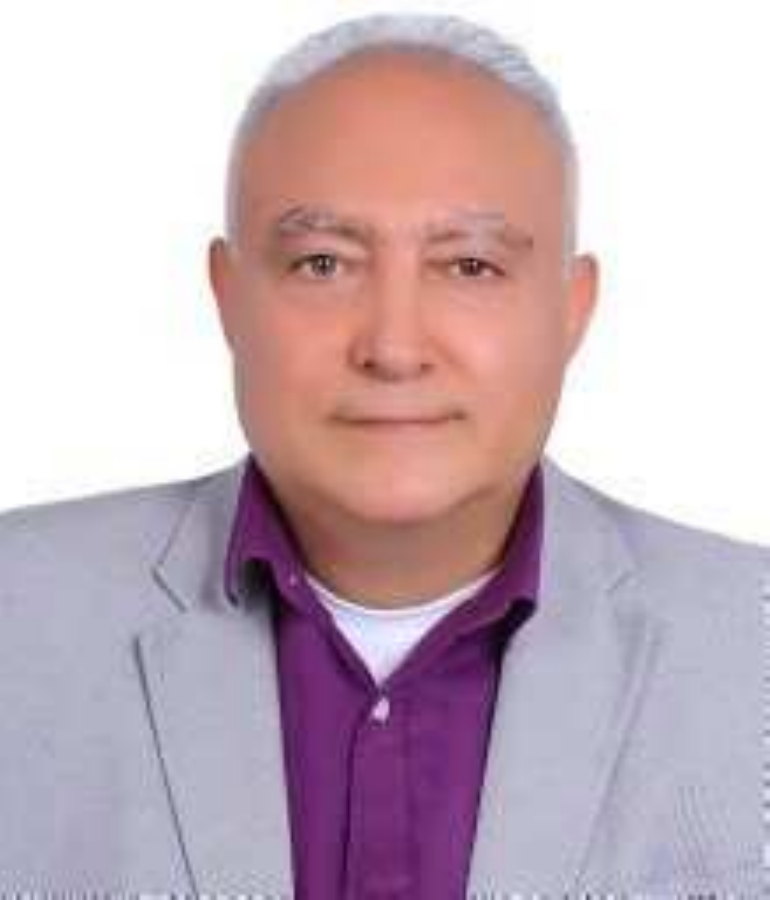}}]{Ali Movaghar}
is a Professor in the Department of Computer Engineering at Sharif University of Technology in Tehran. He received his M.S. and Ph.D. degrees in Computer, Information, and Control Engineering from the University of Michigan, in 1979 and 1985, respectively. His research interests include performance/dependability modeling, distributed real-time systems, and Big Data. He is a senior member of the IEEE and the ACM.
\end{IEEEbiography}
\vspace{-45pt}
\begin{IEEEbiography}[{\includegraphics[width=1in,height=1.1in,clip,keepaspectratio]{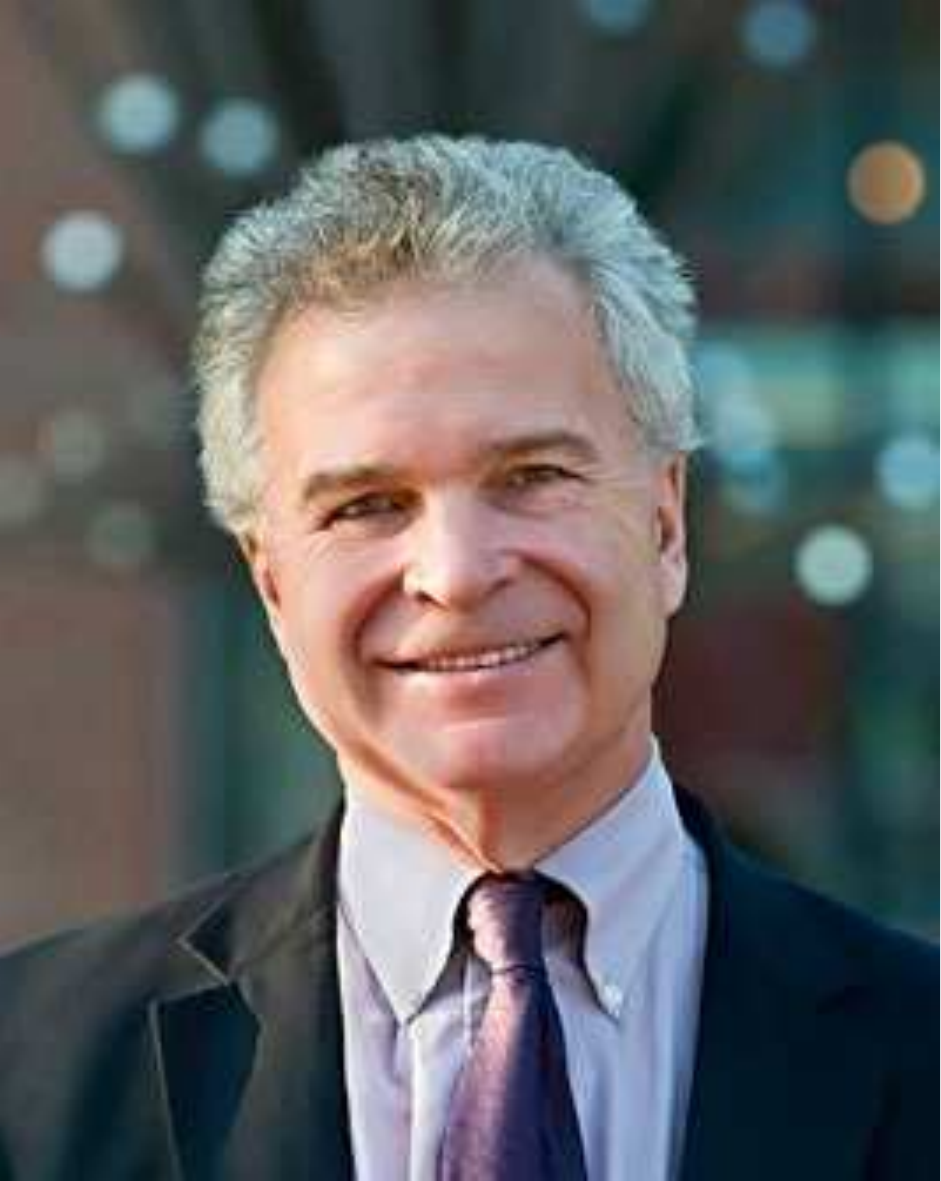}}]{H. Eugene Stanley}
received his Ph.D. degree in physics from Harvard University in 1967. He is an American Physicist and currently a University Professor at Boston University, USA. He has made fundamental contributions to complex systems and is one of the founding fathers of econophysics. His current research interests include complexity science and econometrics. He was elected to the U.S. National Academy of Sciences in 2004.
\end{IEEEbiography}



\end{document}